\ifdefined\XeTeXrevision
\else
\pdfoutput=1
\fi

\documentclass[11pt]{article}
\usepackage{algorithm}
\usepackage{amsmath}
\usepackage{algpseudocode}
\usepackage{colortbl}
\usepackage{xcolor}
\usepackage{pifont}
\usepackage{makecell}
\definecolor{GrayHeader}{gray}{0.9}
\definecolor{CheckGreen}{rgb}{0.1, 0.6, 0.1}
\definecolor{CrossRed}{rgb}{0.8, 0.0, 0.0}
\definecolor{LightBlue}{RGB}{230, 242, 255}
\definecolor{CodeBg}{RGB}{245, 245, 245}

\usepackage[final]{acl}
\usepackage{times}
\usepackage{latexsym}
\usepackage[T1]{fontenc}
\usepackage[utf8]{inputenc}
\usepackage{microtype}
\usepackage{inconsolata}
\usepackage{graphicx}
\usepackage{enumitem}
\usepackage{booktabs}
\usepackage{multirow}
\usepackage{adjustbox}
\usepackage{listings}
\usepackage{subcaption}

\title{MisconceptDialogue: Evaluation Criteria Shape Measured Success in LLM Math Tutoring}

\author{
Woojin Lee$^{1}$\thanks{Co-first authors} \quad
Yoonkyu Woo$^{1,2}$\footnotemark[1] \quad
Jinah Kim$^{1,2}$ \quad
Hyeok-Min Gwon$^{1}$ \quad
Jin-Xia Huang$^{1,2}$\thanks{Corresponding author} \\
\\
$^{1}$Electronics and Telecommunications Research Institute, Republic of Korea \\
$^{2}$Department of Artificial Intelligence, University of Science and Technology, Republic of Korea \\
\texttt{\{writerwoody, yoon303, wlsdk3165, hgha1194990, hgh\}@etri.re.kr}
}

\begin{document}
\maketitle

\begin{abstract}
Large language models are increasingly used as math tutors, raising two questions: can a single LLM detect a student's error, diagnose the underlying misconception, and remediate it through multi-turn dialogue---and can we trust the numbers that report how well it does? We address both with \textsc{MisconceptDialogue}, a fully automated framework that runs the full pipeline on five open-weight LLM tutors (7B--14B) interacting with LLM-simulated students over 5,000+ dialogues---a deliberately simulated protocol (paired conditions, a wrong-diagnosis control unsafe with real learners, scale) whose results do not establish real-student learning. \textbf{On measurement}, our headline finding is uncomfortable for the field: benchmark-style exact-match scoring common in standard math NLP---a lower-bound stress test---reports 42.1\% measured Remediation Success Rate on identical dialogues where a human-calibrated LLM-as-judge reports 81.3\%, a 39.2 percentage-point spread larger than the combined effects of tutor model, teaching strategy, and supplied diagnostic information. This is not a single-judge artifact: four independent LLM judges and 200 human annotations converge to within a few points ($\kappa$ up to 0.91). \textbf{On capability}, the verdict is equally striking: the score barely changes whether we tell the tutor the student's actual misconception or a deliberately swapped one---under the tested diagnosis-presentation format, these tutors show little evidence of reliably conditioning on the diagnostic information they are handed. For LLM tutoring evaluation to mature, multi-criterion scoring with multi-judge convergence and human anchoring should become standard practice---and the measurement question deserves as much scrutiny as the capability question it claims to answer.
\end{abstract}

\section{Introduction}
\label{sec:intro}

Student misconceptions---systematic, coherent patterns of incorrect reasoning---are among the most persistent obstacles to learning. Unlike careless errors, they reflect deep-seated misunderstandings that resist correction through mere presentation of correct answers \cite{vosniadou2004extending}: Conceptual Change Theory \cite{posner1982accommodation} holds that remediation requires \textit{cognitive conflict}, and \citet{chi2005commonsense} shows that misconceptions of different ontological categories demand fundamentally different interventions. Decades of research on intelligent tutoring systems (ITS) have demonstrated that automated instruction can approach the effectiveness of expert human tutoring \cite{vanlehn2011relative, kulik2016effectiveness}, but only when the system accurately diagnoses the misconception and selects an appropriate pedagogical strategy \cite{confrey1990chapter, smith1994misconceptions, tippett2010refutation}.

Large language models are now assuming this role. A randomized controlled trial shows that GPT-4-based tutoring significantly improves student engagement and learning outcomes \cite{vanzo2025gpt}; reinforcement learning enables 7B-scale models to achieve pedagogical alignment competitive with proprietary systems \cite{dinucu2025problem}; and inference-time strategy steering allows explicit control over tutoring behavior \cite{puech2025towards}. Yet rigorous benchmark studies temper this enthusiasm: MathTutorBench \cite{macina2025mathtutorbench} finds substantial gaps between LLM and human tutors, and GuideEval \cite{liu2025discerning} shows that LLMs fail to adapt guidance to learner cognitive states. As LLM tutors are deployed at scale \cite{kasneci2023chatgpt}, two fundamental questions arise---one about \emph{capability}, one about \emph{measurement}.

\textbf{The capability question:} How well does the full \textit{detect--diagnose--remediate} pipeline work? Existing benchmarks evaluate individual stages in isolation---the BEA 2025 Shared Task \cite{kochmar2025findings} measures mistake identification, the EEDI Kaggle 2024 competition \cite{king2024eedi} tests misconception classification from multiple-choice distractors---but the full pipeline remains under-evaluated within multi-turn dialogue. This matters because the three stages are deeply interdependent: incorrect detection leads to missed interventions, misdiagnosis leads to inappropriate strategies, and even correct diagnosis fails if the tutor reveals the answer rather than guiding understanding \cite{macina2025mathtutorbench}.

\textbf{The measurement question:} Can we trust the numbers? In tutoring, a student's final answer string, numerical equivalence, and conceptual understanding are related but distinct targets. Evaluation choices that are routine in standard math NLP may therefore measure different constructs when applied to multi-turn remediation dialogues. At the same time, LLM-as-judge evaluation \cite{zheng2023judging} carries documented biases \cite{panickssery2024llm, chen2024humans}, and LLM-simulated students used for scalable evaluation raise their own ecological validity questions \cite{martynova2025can, wu2025embracing}.

We present \textsc{MisconceptDialogue}, a fully automated evaluation framework that uses LLM-simulated students for scale while decoupling the tutoring pipeline from the scoring lenses used to measure it \cite{scarlatos2026simulated}. Simulation is a deliberate trade of ecological validity for experimental control: it holds the student fixed while tutor-side information varies across six paired pipeline conditions, permits a wrong-diagnosis control that could not responsibly be run on real struggling students, and scales to over 5,000 dialogues so that criterion effects carry paired statistics. Every capability claim below is therefore scoped to this protocol. Through the first integrated evaluation of the complete \textit{detect--diagnose--remediate} pipeline---spanning five open-weight LLM tutors (7B--14B) across four experiments and over 5,000 dialogues---we find that at this scale, the choice of evaluation criterion shapes measured tutoring success more than any modeling lever we manipulate. Our five findings group into measurement (1--2) and capability (3--5):

\begin{enumerate}[itemsep=1pt, parsep=0pt]
\item \textbf{Evaluation-criterion sensitivity}: Applying benchmark-style exact-match scoring common in math NLP---a lower-bound stress test of imported benchmark practice, not a normative tutoring metric---reports 42.1\%\ \textbf{Remediation Success Rate (RSR)}, the fraction of scenarios where the student solves a held-out verification problem after tutoring, on the same dialogues where a human-calibrated LLM-as-judge reports 81.3\%---a 39.2\,pp gap showing that scoring conventions can substantially shift measured tutoring success (\S\ref{sec:exp3}).

\item \textbf{Judge robustness and human calibration}: The gap in (1) is not an artifact of one lenient judge---four LLM judges from three vendor families converge within 6\,pp on the same dialogues (Fleiss $\kappa$=0.85), and on human-annotated subsets they align with human consensus on clean cases ($\kappa$=0.79--0.91) while remaining informative on borderline cases ($\kappa$=0.74--0.87) (\S\ref{sec:human-eval}).

\item \textbf{Context-aware detection}: Turning to capability, even the first pipeline stage proves brittle---multi-turn context improves apparent detection on imbalanced data but \emph{degrades} it on mixed-label data across all five models. The apparent gains were bias exploitation, not skill (\S\ref{sec:exp1}).

\item \textbf{Diagnosis--detection divergence}: One stage up the pipeline, detection strength fails to predict diagnosis accuracy across all five models---exposing a category error in benchmarks that conflate ``mistake identification'' with ``misconception identification'' (\S\ref{sec:exp2}).

\item \textbf{Pipeline null result}: If this decoupling could be patched by handing the tutor a perfect diagnosis, progressive oracle ablation finds no detectable benefit from structured pipeline information over direct tutoring under this protocol ($p$=0.263), and a three-judge wrong-diagnosis control provides little evidence that these tutors reliably condition on the diagnostic labels they are handed under the tested diagnosis-presentation format (\S\ref{sec:exp4}).
\end{enumerate}

\section{Related Work}
\label{sec:related}

\subsection{From Detection to Diagnosis}
Work on misconception-aware math tutoring has developed along two mostly separate lines. One diagnoses misconceptions from structured artifacts: rule-based diagnosis systems \cite{feldman2018automatic}, LLM-based error classification \cite{mcnichols2023algebra}, and distractor-based retrieval in EEDI \cite{wang2021results,king2024eedi}. The other moves into dialogue, where BEA 2025 scores mistake identification and feedback quality (\citealp{kochmar2025findings}). This split matters for our capability question: detecting that a response is wrong is not the same as identifying the misconception a tutor must remediate. MathDial \cite{macina2023mathdial}, Bridge \cite{wang2024bridging}, SocraticLM \cite{liu2024socraticlm}, ConvoLearn \cite{sharma2026convolearn}, and the Math Misconceptions Benchmark \cite{otero2025benchmark} make this dependency testable in context, but whether contextual detection predicts fine-grained diagnosis within one tutoring pipeline remains open.

\subsection{LLM Tutoring and the End-to-End Evaluation Gap}
Dialogue-based tutoring has expanded from AutoTutor \cite{graesser2004autotutor, nye2014autotutor} and early LLM evaluations \cite{tack2022ai} to Socratic, RL-aligned, strategy-steered, fine-tuned, and evaluation-focused LLM tutoring systems \cite{peng2025kele,dinucu2025problem,puech2025towards,team2024learnlm,shi2026solver,hsu2026mathedu}. Yet \citet{borchers2025can} show LLMs still trail traditional ITS in adaptivity, and MathTutorBench \cite{macina2025mathtutorbench} documents persistent gaps versus human tutors. Prior evaluations are component-centered, leaving it under-tested whether the detect$\to$diagnose$\to$remediate structure improves over direct tutoring---a learning-science-motivated hypothesis \citealp{chi2005commonsense,tippett2010refutation} rarely tested as an integrated pipeline.

\subsection{The Measurement Problem}
\label{sec:rw-measurement}
The measurement question (\S\ref{sec:intro}) arises from two converging concerns.
First, the LLM-as-judge paradigm \cite{zheng2023judging} that dominates open-ended NLP evaluation carries well-documented biases: models favor their own outputs \cite{panickssery2024llm}, and both human and LLM judges are vulnerable to misinformation oversight, authority, and gender biases under controlled perturbations \cite{chen2024humans}. For tutoring evaluation---where ``correct understanding'' is inherently more ambiguous than factual verification---these biases could systematically distort measured remediation success, yet their impact remains underexplored. Second, the LLM-simulated students increasingly used as scalable proxies for real learners \cite{markel2023gpteach, scarlatos2026simulated} introduce their own validity concerns: teachers report significant gaps in authenticity and emotional range \cite{martynova2025can}, and faithfully modeling diverse cognitive levels remains an open challenge \cite{wu2025embracing}. How closely simulated learners track real ones is itself an active measurement question, approached through Turing-like tests for simulated learners \cite{kaser2024simulated} and psychometric alignment with human knowledge distributions \cite{heyueya2024psychometric}.

\paragraph{Synthesis.} Together, these gaps leave it unclear how much measured RSR reflects tutoring behavior rather than evaluation artifacts. \S\ref{sec:framework} introduces \textsc{MisconceptDialogue} to address them jointly.

\section{The MisconceptDialogue Framework}
\label{sec:framework}

\begin{figure}[t]
\centering
\includegraphics[width=\columnwidth]{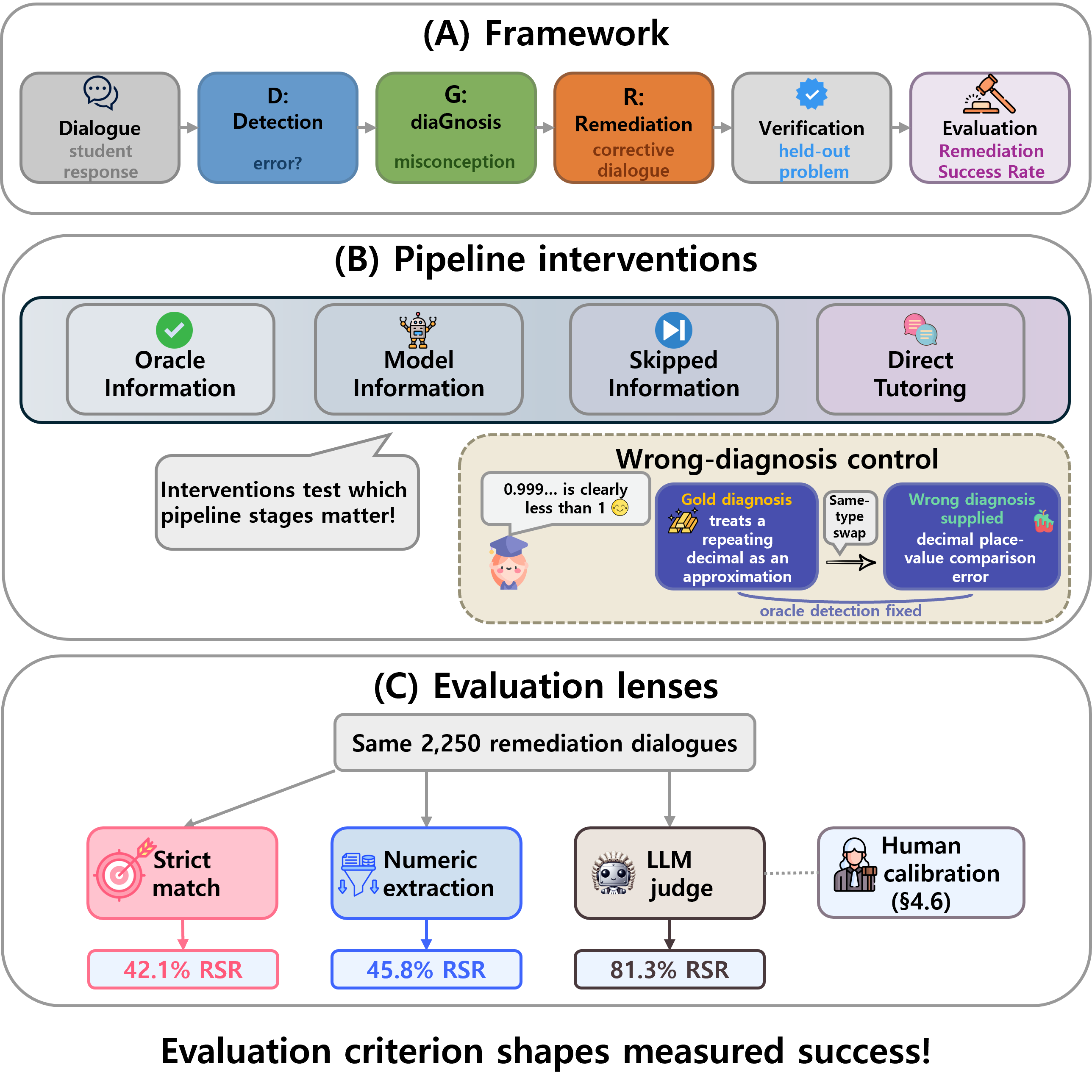}
\caption{Overview of \textsc{MisconceptDialogue}. \textbf{(A)} The D-G-R pipeline moves from student dialogue through detection, diagnosis, remediation, verification, and final evaluation. \textbf{(B)} Pipeline-information interventions vary what information the tutor receives---oracle, model-generated, skipped, or direct tutoring---with a wrong-diagnosis control that swaps the diagnosis while holding detection fixed (\S\ref{sec:exp4}). \textbf{(C)} Three evaluation lenses---strict match, numeric extraction, and a human-calibrated LLM-as-judge (\S\ref{sec:human-eval})---score the same 2,250 remediation dialogues, shifting measured RSR from 42.1\% to 81.3\%.}
\label{fig:pipeline}
\end{figure}

\textsc{MisconceptDialogue} separates three objects that are often conflated: (i) the D-G-R tutoring pipeline, (ii) pipeline-information interventions controlling what the tutor is told, and (iii) evaluation lenses applied to the same resulting dialogues (Figure~\ref*{fig:pipeline}). This separation lets Exp.~3 ask how scoring criteria change measured remediation, and Exp.~4 ask whether diagnostic information changes tutor behavior.

\paragraph{Misconception taxonomy.}
\label{sec:taxonomy}
We adopt 55 educator-validated algebra misconceptions from the Math Misconceptions Benchmark \cite{otero2025benchmark}, categorized by \textit{type} (factual, procedural, or conceptual) and \textit{domain} (fractions, algebra, geometry, arithmetic, number theory). This type distinction is pedagogically meaningful: different misconception categories can call for different interventions \cite{chi2005commonsense}. Exp.~2 diagnosis uses these 55 entries as a closed-set taxonomy. For Exp.~3/4 remediation, we instantiate them into 106 evaluation scenarios by varying the problem context (36 factual, 34 procedural, 36 conceptual). Each scenario carries (i) a problem statement, (ii) an expected student response exhibiting the target misconception, (iii) a free-text label paraphrasing the misconception within the 55-class taxonomy, and (iv) a \textbf{held-out verification problem} (with gold answer) posed to the student \emph{after} tutoring to determine RSR. Scenarios mix existing benchmark items (53\%) and GPT-5.4-generated variants (47\%).

\paragraph{Remediation strategies.}
\label{sec:strategies}
Drawing on learning science research \cite{posner1982accommodation, tippett2010refutation}, we define five strategies: \textbf{Counterexample} (cases where the misconception fails), \textbf{Analogy} (links to familiar concepts), \textbf{Cognitive Conflict} (contradictions in student reasoning), \textbf{Direct Instruction} (explicit rule explanation), and \textbf{Socratic Questioning} (guided discovery). Exp.~3 fixes one strategy to isolate effects; Exp.~4's free-strategy condition lets the tutor choose from the diagnosis (\S\ref{sec:exp3}, \S\ref{sec:exp4}).

\paragraph{Evaluation metrics.}
Our central methodological move is to score \emph{identical dialogues with multiple criteria}, decomposing measured RSR into what the tutor produces versus what the scorer credits (\S\ref{sec:intro}). For remediation we report RSR---the fraction of scenarios where the simulated student solves a held-out verification problem after tutoring (\S\ref{sec:intro})---under three criteria: (a) benchmark-style exact-match regex common in answer-focused math NLP, (b) numerical extraction with tolerance, and (c) a human-calibrated LLM-as-judge (\S\ref{sec:human-eval}). We run the LLM judge with four models from three vendor families (GPT-5-mini and GPT-5.4 \cite{singh2025openai}, Claude Haiku 4.5 \cite{anthropic2025haiku}, Gemini 3.1 Flash Lite \cite{gdm2026gemini}; \S\ref{sec:exp3}); Claude Haiku 4.5 and Gemini 3.1 Flash Lite also serve as out-of-family tutors in \S\ref{sec:exp5}. We additionally report \textbf{Answer Reveal Rate (ARR)}---the fraction of dialogues where the tutor reveals the target answer during tutoring. Other stages use standard component metrics: three-class Macro F1 for detection \cite{kochmar2025findings} and Accuracy@1 for diagnosis. All prompts are specified in Appendix~\ref{sec:appendix-prompts}, simulated-student profiles in Appendix~\ref{sec:appendix-profiles}, and decoding and run-level hyperparameters in Appendix~\ref{sec:appendix-hyperparams}.

\section{Experiments}
\label{sec:experiments}

\subsection{Models and Datasets}
We evaluate five open-weight models in the 7B--14B range: Qwen-2.5-7B \cite{qwen2024qwen25technicalreport}, Mistral-7B \cite{jiang2023mistral7b}, Llama-3.1-8B \cite{grattafiori2024llama}, Phi-4 \cite{abdin2024phi} (14B), and Qwen3.5-9B \cite{yang2025qwen3technicalreport}\footnote{\url{https://huggingface.co/Qwen/Qwen3.5-9B}} (9B). The five local models run in bfloat16 on H100 GPUs (Appendix~\ref{sec:appendix-reproducibility}). For detection (Exp.~1), we use three datasets: BEA 2025 \cite{kochmar2025findings} ($n$=300, imbalanced), MathDial \cite{macina2023mathdial} ($n$=2{,}262, 75\%/25\% correct/incorrect), and Bridge \cite{wang2024bridging} ($n$=419, imbalanced). For diagnosis (Exp.~2), we use the Math Misconceptions Benchmark \cite{otero2025benchmark} (55 misconceptions, 220 examples). Per-dataset statistics and label distributions are in Appendices~\ref{sec:appendix-data} and~\ref{sec:appendix-labels}. For the out-of-family replication (\S\ref{sec:exp5}), we additionally evaluate Claude Haiku 4.5 and Gemini 3.1 Flash Lite as tutors on the same protocol.

\subsection{Experiment 1: Context-Aware Detection}
\label{sec:exp1}

\paragraph{Setup.}
Following BEA 2025 Track 1, we evaluate mistake identification under three context conditions: \textbf{single-turn} (final response only), \textbf{multi-turn~(2)} (two prior turns), and \textbf{multi-turn~(full)} (complete history).

\paragraph{Results.}
A sharp dichotomy emerges along dataset balance (Table~\ref{tab:exp1}). On imbalanced datasets (BEA 2025 and Bridge, containing \emph{only} incorrect responses), multi-turn context improves F1 for all five models. On mixed-label MathDial (75\%/25\% correct/incorrect), the same five models all \emph{lose} F1 with full context ($\Delta$: $-$0.026 to $-$0.163). This consistent pattern indicates that context-driven gains on imbalanced data reflect class-prior exploitation rather than better discrimination. Subsampling itself can mislead: Qwen3.5-9B's positive context effect at $n$=300 ($\Delta$=+0.101) reverses to $-$0.163 on the full dataset.

A label-shift mechanism underlies the effect: predictions skew toward ``incorrect'' as context grows (mean 55.8\%$\rightarrow$79.7\%), correct-class recall collapses (0.444$\rightarrow$0.202), and incorrect-class recall rises (0.565$\rightarrow$0.793). On datasets where every instance is incorrect this shift inflates aggregate scores; on MathDial it degrades F1 by misclassifying correct responses.

\begin{table*}[t]
\centering
\small
\adjustbox{max width=\textwidth}{%
\begin{tabular}{llcccccccccc}
\toprule
& & \multicolumn{2}{c}{\textbf{Qwen-2.5-7B}} & \multicolumn{2}{c}{\textbf{Mistral-7B}} & \multicolumn{2}{c}{\textbf{Llama-3.1-8B}} & \multicolumn{2}{c}{\textbf{Phi-4}} & \multicolumn{2}{c}{\textbf{Qwen3.5-9B}} \\
\cmidrule(lr){3-4} \cmidrule(lr){5-6} \cmidrule(lr){7-8} \cmidrule(lr){9-10} \cmidrule(lr){11-12}
\textbf{Dataset} & \textbf{Context} & \textbf{F1} & \textbf{Acc} & \textbf{F1} & \textbf{Acc} & \textbf{F1} & \textbf{Acc} & \textbf{F1} & \textbf{Acc} & \textbf{F1} & \textbf{Acc} \\
\midrule
\multirow{3}{*}{BEA 2025} & Single & 0.237 & 0.553 & 0.108 & 0.193 & 0.253 & 0.610 & 0.257 & 0.627 & 0.218 & 0.487 \\
& Multi (2) & 0.259 & 0.633 & 0.173 & 0.350 & 0.276 & 0.707 & 0.284 & 0.743 & 0.234 & 0.540 \\
& Multi (full) & \textbf{0.309} & \textbf{0.863} & \textbf{0.187} & \textbf{0.390} & \textbf{0.290} & \textbf{0.770} & \textbf{0.307} & \textbf{0.853} & \textbf{0.285} & \textbf{0.747} \\
\midrule
\multirow{3}{*}{MathDial} & Single & \textbf{0.473} & \textbf{0.509} & \textbf{0.492} & \textbf{0.630} & \textbf{0.430} & \textbf{0.442} & \textbf{0.345} & \textbf{0.348} & \textbf{0.438} & \textbf{0.445} \\
& Multi (2) & 0.426 & 0.435 & 0.439 & 0.468 & 0.365 & 0.365 & 0.214 & 0.325 & 0.402 & 0.404 \\
& Multi (full) & 0.314 & 0.320 & 0.466 & 0.499 & 0.327 & 0.335 & 0.289 & 0.302 & 0.275 & 0.292 \\
\midrule
\multirow{3}{*}{Bridge} & Single & 0.204 & 0.442 & 0.131 & 0.150 & 0.158 & 0.463 & 0.011 & 0.017 & 0.014 & 0.021 \\
& Multi (2) & 0.263 & 0.652 & 0.203 & 0.439 & 0.242 & 0.570 & 0.119 & 0.217 & 0.097 & 0.169 \\
& Multi (full) & \textbf{0.301} & \textbf{0.823} & \textbf{0.219} & \textbf{0.489} & \textbf{0.257} & \textbf{0.628} & \textbf{0.164} & \textbf{0.327} & \textbf{0.135} & \textbf{0.253} \\
\bottomrule
\end{tabular}%
}
\caption{Exp.~1: Mistake identification (Macro F1 / Accuracy) across five models, three datasets, and three context conditions.}
\label{tab:exp1}
\end{table*}

\subsection{Experiment 2: Diagnosis Accuracy}
\label{sec:exp2}
Experiment~1 addressed whether models detect mistakes; Experiment~2 turns to diagnosis---identifying \emph{which} misconception underlies an error.

\paragraph{Setup.}
On the 220-item Math Misconceptions Benchmark, we compare five 55-way closed-set diagnosis conditions: (A)~\textbf{Zero-shot}, (B)~\textbf{Few-shot} (5 examples), (C)~\textbf{CoT + rubric} \cite{wei2022cot}, (D)~\textbf{RAG} using BGE embeddings \cite{xiao2024c, lewis2020retrieval}, and (E)~\textbf{Generate-Retrieve-Rerank} (GRR) \cite{mitton2026genretrieve}. Appendix~\ref{sec:appendix-diagnosis} details the condition implementations and the post-hoc parsing corrections.

\begin{table}[t]
\centering
\small
\adjustbox{max width=\columnwidth}{%
\begin{tabular}{lccccc}
\toprule
\textbf{Condition} & \textbf{Qwen2.5} & \textbf{Mistral} & \textbf{Llama3.1} & \textbf{Phi-4} & \textbf{Qwen3.5} \\
\midrule
(A) Zero-shot & 0.359 & 0.077 & 0.291 & 0.436 & \textbf{0.545} \\
(B) Few-shot (5) & 0.291 & 0.241 & 0.273 & 0.441 & 0.473 \\
(C) CoT + rubric & 0.355 & 0.214 & 0.286 & 0.423 & 0.509 \\
(D) RAG & 0.355 & 0.232 & 0.236 & 0.400 & 0.459 \\
(E) Gen-Ret-Rerank & \textbf{0.377} & \textbf{0.255} & \textbf{0.332} & \textbf{0.477} & 0.518 \\
\bottomrule
\end{tabular}%
}
\caption{Exp.~2: Misconception diagnosis Acc@1 across five models and five prompting conditions (55 misconceptions, $n$=220). Bold marks the per-model best. Post-hoc parsing corrections: Appendix Table~\ref{tab:parsing-correction}.}
\label{tab:exp2}
\end{table}

\paragraph{Results.}
The headline pattern is that \textbf{detection and diagnosis diverge} (Table~\ref{tab:exp2}): Mistral-7B has the strongest mixed-label detection F1 in Exp.~1 (0.492) yet the lowest best-case diagnosis accuracy ($\leq$0.255), while Qwen3.5-9B leads diagnosis (0.545) despite only moderate detection (0.438)---identifying \emph{whether} a mistake exists and \emph{which} misconception underlies it draw on different capabilities. Retrieval-based GRR lifts the weakest model (Mistral, 0.077$\rightarrow$0.255) but remains below zero-shot for the strongest model (Qwen3.5 zero-shot 0.545 vs.\ GRR 0.518); few-shot is inconsistent ($+$0.002 mean), and procedural cases score highest while conceptual cases score lowest (Appendix~\ref{sec:appendix-bytype}).

\subsection{Experiment 3: Remediation and Evaluation Methodology}
\label{sec:exp3}
Experiments~1--2 addressed detection and diagnosis capability. Experiment~3 turns to the measurement question: does the choice of evaluation criterion materially affect measured remediation success?

\paragraph{Setup.}
We evaluate five tutor models on a 5-strategy $\times$ 3-type matrix over a 106-scenario bank (36 factual, 34 procedural, 36 conceptual). Tiled-shuffle allocation assigns 30 scenarios per strategy--type cell, reusing scenarios across cells to preserve type balance and within-cell diversity, yielding 450 dialogues per model (2,250 total). The tutor follows the assigned strategy; a GPT-5-mini \cite{singh2025openai} simulated student completes 5 tutoring turns and a verification problem.

Crucially, we apply \textbf{three evaluation criteria} to the \emph{same} dialogues: \textbf{Strict match}: word-boundary regex matching---strictest; \textbf{Numeric extraction}: numerical answer extraction with tolerance; \textbf{LLM-as-judge}: GPT-5-mini accepting paraphrases and reasoning evidence---most lenient. Strict match is not an arbitrary baseline: it mirrors the exact-match scoring that standard math NLP benchmarks apply to model outputs, so including it directly tests \emph{how much current benchmark practice would lower measured remediation estimates if transferred to tutoring dialogues}. LLM-as-judge sits at the other end---a criterion explicitly designed to credit understanding beyond exact-answer matching, and the one we calibrate against human consensus (\S\ref{sec:human-eval}).

\paragraph{Results.}
The same dialogues yield systematically different RSR depending on the scoring criterion (Table~\ref{tab:exp3}): Strict match 40.2--44.4\%, Numeric extraction 44.4--47.1\%, LLM judge 77.8--84.4\%. Because the dialogues are identical, the 39.2\,pp gap reflects \emph{what counts as correct}, not changes in tutor behavior or answer revealing (ARR = 6.0--8.7\%). Strict and Numeric are nearly equivalent (42.1\% vs.\ 45.8\%); only the LLM judge is substantially higher, so the gap stems from whether \emph{conceptual understanding} is credited beyond exact-answer matching. Paired McNemar testing on identical dialogues confirms the gap (Strict vs.\ LLM judge: $\chi^2$=765.7, $p<10^{-50}$; 95\% bootstrap CI: [36.6, 41.8]\,pp); Appendix~\ref{sec:appendix-mcnemar} states why a paired test is the appropriate choice here.

\begin{table}[t]
\centering
\small
\adjustbox{max width=\columnwidth}{%
\begin{tabular}{lcccccc}
\toprule
& \multicolumn{2}{c}{\textbf{Automatic criteria}} & \multicolumn{3}{c}{\textbf{LLM-as-judge}} & \\
\cmidrule(lr){2-3} \cmidrule(lr){4-6}
\textbf{Tutor Model} & \textbf{Strict} & \textbf{Numeric} & \textbf{GPT} & \textbf{Claude} & \textbf{Gemini} & \textbf{ARR} \\
 & \textbf{(\%)} & \textbf{(\%)} & \textbf{(\%)} & \textbf{(\%)} & \textbf{(\%)} & \textbf{(\%)} \\
\midrule
Qwen-2.5-7B & 44.4 & 46.4 & 81.8 & 76.2 & 79.1 & 6.9 \\
Phi-4 & 42.4 & 47.1 & 83.6 & 78.2 & 81.1 & 6.4 \\
Qwen3.5-9B & 43.3 & 46.4 & 84.4 & 77.8 & 80.4 & 8.7 \\
Llama-3.1-8B & 40.2 & 44.4 & 79.1 & 74.4 & 76.9 & 6.7 \\
Mistral-7B & 40.2 & 44.4 & 77.8 & 70.2 & 73.6 & 6.0 \\
\midrule
\textbf{Mean} & \textbf{42.1} & \textbf{45.8} & \textbf{81.3} & \textbf{75.4} & \textbf{78.2} & \textbf{6.9} \\
\bottomrule
\end{tabular}%
}
\caption{Exp.~3: RSR (\%) on the same 2,250 dialogues (5 models $\times$ 5 strategies $\times$ 3 types $\times$ 30 scenarios; $n$=450 per row). Columns group Strict/Numeric as automatic criteria and GPT/Claude/Gemini as LLM-as-judge variants: GPT-5-mini, Claude Haiku 4.5, and Gemini 3.1 Flash Lite. GPT-5.4 cross-check is reported in Appendix~\ref{sec:appendix-crossfamily}. ARR = answer reveal rate.}
\label{tab:exp3}
\end{table}

\paragraph{Judge-family convergence.}
To check that the LLM-judge result is not a single-judge artifact, we rescore the same 2,250 dialogues with two non-OpenAI judges---Claude Haiku 4.5 (Anthropic) and Gemini 3.1 Flash Lite (Google)---plus a GPT-5.4 cross-check. Agreement is high: the three main judges (corrected GPT-5-mini, Claude Haiku 4.5, Gemini 3.1 Flash Lite) reach Fleiss $\kappa$=0.85, with pairwise Cohen $\kappa$=0.78--0.93. Per-model RSR rankings (Mistral-7B lowest, Phi-4/Qwen3.5-9B highest) survive across judges. A 3-judge majority-vote RSR of 78.2\% sits within 3\,pp of each individual judge, and GPT-5.4 tracks GPT-5-mini within 2.8\,pp under aligned evaluation context (Appendix~\ref{sec:appendix-crossfamily}). The 39.2\,pp gap in Table~\ref{tab:exp3} is therefore a property of the criterion, not of any specific judge.

\paragraph{Disagreement taxonomy.} Cross-tabulating the three criteria on the same 2,250 dialogues decomposes the 39.2\,pp gap into two distinct components. Strict-fail/Judge-pass disagreements total 949 (42.2\% of all dialogues): \textbf{23.4\% (222) are also Numeric-pass}, exposing string-mismatches on numerically correct answers; \textbf{76.6\% (727) are Numeric-fail/Judge-pass}, capturing paraphrase tolerance and reasoning credit beyond exact-answer matching. Roughly three-quarters of the gap is therefore construct-credit difference, with the remaining quarter attributable to the surface-matching brittleness of benchmark-style exact matching. Both components support the same conclusion: current benchmark defaults should not transfer wholesale to tutoring evaluation.

Non-measurement effects split between a strong type effect and a flat strategy effect. The LLM judge shows a clear monotone ordering by misconception type (conceptual 89.9\% > procedural 80.9\% > factual 73.2\%; consistent across all five models, $p<10^{-30}$), an ordering that nearly vanishes under strict match---the type effect surfaces through conceptual-understanding scoring (Appendix~\ref{sec:appendix-bytype}). Strategy effects are essentially flat: the five strategies cluster within 6.7\,pp under LLM-judge RSR (averaged over models) with no significant strategy$\times$type interaction; Appendix~\ref{sec:appendix-strategy-examples} gives dialogue excerpts illustrating this invariance. Two of the five (Direct Instruction, Counterexample) have low self-marker ratios (<\,0.11; Appendix~\ref{sec:appendix-fidelity}), so part of the apparent equivalence may reflect imperfect strategy operationalization; restricting to the three high-fidelity strategies preserves the 6.7\,pp range, so the equivalence is not purely an artifact. The simulated setting nonetheless bounds rather than establishes pedagogical equivalence.

The simulated student adopts the assigned misconception faithfully (persona-only RSR of 6.6\%) and recovers under tutoring in a type-dependent way (conceptual gains exceed procedural by 25.6\,pp), arguing against a simple ``persuadability ceiling'' explanation for the flat strategy effects (Appendix~\ref{sec:appendix-persona}).

\subsection{Experiment 4: End-to-End Pipeline}
\label{sec:exp4}

\paragraph{Setup.}
To test capability directly, we run progressive oracle ablation across five models and the same 106 scenarios, using GPT-5-mini \cite{singh2025openai} as simulated student/judge (3,180 dialogues). Conditions are \textbf{C1}~(gold detection + diagnosis + Exp.~3 type-best strategy\footnote{C1 maps factual/conceptual$\to$direct instruction and procedural$\to$socratic; for procedural and conceptual the top-2 strategies differ by $<$7\,pp (n.s.), so this mapping is approximate.}), \textbf{C2}~(oracle detection + diagnosis, free strategy), \textbf{C3}~(oracle detection + model diagnosis), \textbf{C3.5}~(oracle detection only, no diagnosis), \textbf{C4}~(fully automated pipeline), and \textbf{B1}~(direct tutoring, no pipeline).

\paragraph{Results.}
RSR shows no significant pipeline-condition effect (GLMM $\chi^2$=6.47, $p$=0.263, Table~\ref{tab:exp4}). The RSR range from 76.6\% (C2) to 81.3\% (C3) yields no significant pairwise differences (Holm-corrected $p$=1.000, Cohen's $h<0.10$). TOST equivalence testing \cite{schuirmann1987comparison} accepts $\delta$=10\,pp for all 15 pairs, but only three near-identical pairs at $\delta$=5\,pp, so pipeline benefits around 5\,pp remain possible for most comparisons (Appendix~\ref{sec:appendix-tost}). The pattern is criterion-invariant: strict match shows a similar 3.1\,pp condition spread versus the LLM judge's 4.9\,pp. At $n$=530 per condition, the design has 80\% power for effects $\geq$7\,pp but only 56\% at 5\,pp. The absolute RSR (74--78\%) tracks Exp.~3's LLM-judge RSR (78--84\%).

A more direct control reaches the same conclusion: supplying oracle detection with an intentionally mismatched same-type misconception yields RSR of 80.8\%---statistically indistinguishable from gold diagnosis (C2, 76.6\%; McNemar $p$=0.078) or direct tutoring (B1, 78.5\%; $p$=0.315). Across the same 1,590 dialogues, three independent judges agree on the wrong-vs-gold direction (Claude Haiku 4.5 +3.2\,pp, Gemini 3.1 Flash Lite +3.4\,pp, GPT-5-mini +4.2\,pp), arguing against a single-judge artifact. Four readings remain compatible: (i)~pipeline structure adds little detectable signal under this protocol; (ii)~7B--14B tutors fail to condition on the diagnostic label; (iii)~the simulated student recovers from generic tutoring regardless of diagnostic correctness; (iv)~the LLM judge cannot discriminate the differences. Because the swap holds coarse misconception type fixed, the control targets fine-grained label uptake rather than cross-type diagnostic usefulness. The 3-judge convergence favors readings (ii)--(iii) over a single-judge artifact; (iii) is bounded by the simulator's persona-only RSR of 6.6\% and 21.2\,pp profile-level variance, showing it is not a uniform recoverer (Appendix~\ref{sec:appendix-wrong-diag}).

\begin{table}[t]
\centering
\small
\adjustbox{max width=\columnwidth}{%
\begin{tabular}{lccccc|c}
\toprule
\textbf{Condition} & \textbf{Qwen-2.5} & \textbf{Mistral} & \textbf{Llama-3.1} & \textbf{Phi-4} & \textbf{Qwen3.5} & \textbf{Mean} \\
\midrule
C1: Oracle All & .783 & .726 & .802 & .868 & .849 & .806 \\
C2: Oracle Diag. & .783 & .689 & .755 & .811 & .792 & .766 \\
C3: Model Diag. & .764 & .792 & .764 & .906 & .840 & .813 \\
C3.5: Det. Only & .774 & .755 & .736 & .811 & .821 & .779 \\
C4: Auto Pipeline & .755 & .764 & .802 & .792 & .840 & .791 \\
B1: Direct & .811 & .726 & .774 & .821 & .792 & .785 \\
\bottomrule
\end{tabular}%
}
\caption{Exp.~4: Progressive oracle ablation (RSR). C1=oracle all/type-best strategy, C2=oracle diagnosis/free strategy, C3=model diagnosis, C3.5=detection only, C4=automated pipeline, B1=direct tutoring. Five models $\times$ six conditions $\times$ 106 scenarios, GPT-5-mini evaluator. No condition effect is significant (statistics in \S\ref{sec:exp4}).}
\label{tab:exp4}
\end{table}

\paragraph{Analysis.}
Variance decomposition explains the null: scenario difficulty dominates (intraclass correlation, ICC=0.206), while model and condition effects are negligible (ICC$<$1\%), so pipeline-information signals are smaller than scenario-level variation in this design. Measurable effects are pedagogical rather than procedural: misconception type drives the main effect ($p<10^{-30}$), with no condition$\times$type interaction, and C2 free-strategy collapses into \emph{strategy monoculture} (Appendix~\ref{sec:appendix-fidelity}), recording the lowest mean RSR (76.6\%). C4 detection reaches 96.8\%, making detection an unlikely bottleneck.

\paragraph{Out-of-family tutor replication.}
\label{sec:exp5}
Two concerns about the Exp.~4 null call for an out-of-family check: our student simulator and primary judges share an OpenAI lineage (potential in-family coupling), and our tutors all sit in the 7B--14B band (potential capability saturation). We address both by replicating Exp.~3 with two stronger proprietary tutors from non-OpenAI families---Claude Haiku 4.5 (Anthropic) and Gemini 3.1 Flash Lite (Google)---under the same protocol (Appendix~\ref{sec:appendix-frontier}). Strict-match RSR is essentially unchanged from the 5-baseline mean (43.1\% vs.\ 42.1\%, +1.0\,pp), so the 39.2\,pp criterion gap from \S\ref{sec:exp3} persists at this capability scale. Under the LLM judges, the non-OpenAI mean exceeds the baseline mean by 4.8--6.5\,pp but stays within the 4-judge envelope (judge spread 5.9\,pp), and per-strategy spreads remain narrow (Claude Haiku 4.5 11\,pp, Gemini 3.1 Flash Lite 9\,pp; baseline 6.7\,pp). Moving to stronger out-of-family tutors changes LLM-judge RSR by only single-digit points while the measurement criterion moves it by 40+\,pp---reducing concerns that the main pattern is solely an in-family or 7B--14B saturation artifact.

\subsection{Human Evaluation}
\label{sec:human-eval}

\paragraph{Setup.}
Two quality-stratified subsets calibrate the automated scores against human judgment: a clean set ($n$=100) and a borderline set ($n$=100). Three judge-blinded annotators, each with at least a bachelor's degree and prior coursework or equivalent experience in benchmark-level mathematics, rate each dialogue on four dimensions---correct understanding, reasoning quality, answer revelation, and naturalness---with sampling pipeline, rubric, and annotator details in Appendix~\ref{sec:appendix-humaneval-detail}.

\begin{table}[t]
\centering
\small
\setlength{\tabcolsep}{3pt}
\begin{adjustbox}{max width=\columnwidth}
\begin{tabular}{llrcc}
\toprule
\textbf{Subset} & \textbf{Evaluator} & \textbf{RSR} & \textbf{$\kappa$ vs.\ H} & \textbf{Agree} \\
\midrule
\multirow{6}{*}{\textit{Clean}} & Human (consensus) & 88\% & --- & --- \\
& GPT-5-mini judge & 89\% & 0.853 & 97\% \\
& GPT-5.4 judge & 87\% & 0.863 & 97\% \\
& Claude Haiku 4.5 & 85\% & 0.786 & 95\% \\
& Gemini 3.1 Flash Lite & \textbf{88\%} & \textbf{0.905} & \textbf{98\%} \\
& Strict match & 52\% & 0.175 & 60\% \\
\midrule
\multirow{5}{*}{\textit{Borderline}} & Human (consensus) & 75\% & --- & --- \\
& GPT-5-mini judge & 73\% & 0.740 & 90\% \\
& GPT-5.4 judge & 70\% & 0.775 & 91\% \\
& Claude Haiku 4.5 & 68\% & 0.829 & 93\% \\
& Gemini 3.1 Flash Lite & \textbf{72\%} & \textbf{0.872} & \textbf{95\%} \\
\bottomrule
\end{tabular}
\end{adjustbox}
\caption{Human vs.\ automated evaluation on clean ($n$=100, quality $\geq$3.0) and borderline ($n$=100, quality 2.0--2.99) subsets. Human RSR is three-annotator majority vote; inter-annotator agreement (avg.\ pairwise Cohen's $\kappa$): 0.51 clean, 0.55 borderline. $\kappa$ vs.\ H = evaluator's Cohen's $\kappa$ against human consensus.}
\label{tab:human-eval}
\end{table}

\paragraph{Results.}
Inter-annotator agreement is moderate (mean pairwise Cohen's $\kappa$=0.51 clean, 0.55 borderline; Table~\ref{tab:human-eval}). The four LLM judges track human consensus closely---within 1--3\,pp on clean cases and 2--7\,pp on borderline cases---with item-level Cohen $\kappa$ ranging 0.79--0.91 on clean and 0.74--0.87 on borderline. The paper-default GPT-5-mini reaches $\kappa$=0.853 on clean (97\% agreement; three disagreements analyzed in Appendix~\ref{sec:appendix-disagreements}) and $\kappa$=0.740 on borderline (90\%). The highest agreement on both subsets comes from Gemini 3.1 Flash Lite---a Google model with no OpenAI lineage ($\kappa$=0.905 clean, 0.872 borderline)---showing that judge reliability on coherent dialogues is not specific to any vendor family. Because all four judges track human consensus closely, the near-zero cross-family ICC in \S\ref{sec:exp4} reflects differing evaluator standards rather than a weak underlying signal.

\section{Discussion}
\label{sec:analysis}

\textbf{On measurement}, our two questions answer asymmetrically. The strict-match-to-LLM-judge gap (42.1--81.3\%) on identical dialogues exceeds every other variable: criterion alone shifts RSR by 39.2\,pp while model, strategy, pipeline condition, and judge identity each contribute only single-digit effects. Two non-OpenAI judges reproduce the same gap (33.3--39.2\,pp), the three main judges converge at Fleiss $\kappa$=0.85, and GPT-5.4 tracks GPT-5-mini within 2.8\,pp. The gap therefore lies in the \emph{criterion family}---surface matching versus understanding judgment---rather than any specific judge. We do not treat strict match as a normative tutoring metric; it is a stress test for importing benchmark-style scoring into dialogue, and even numerical extraction remains far below LLM judgment. Item-level agreement between LLM judges and human consensus on clean dialogues ($\kappa$ up to 0.91) supports the understanding-judgment interpretation, so every RSR figure is meaningful \textbf{only relative to its evaluation criterion}.

\textbf{On capability}, perfect pipeline information does not detectably improve remediation under this protocol ($p$=0.263). Because the wrong-diagnosis control holds misconception \emph{type} fixed, this tests fine-grained label uptake rather than rejecting the value of diagnosis wholesale. Several readings remain open: the tutor may not use the label, the simulated student may recover regardless of the diagnosis, or the judge may be insensitive to the difference. What the null is \emph{not} is a simple persuadability ceiling---profile-level RSR varies by 21.2\,pp (Appendix~\ref{sec:appendix-persona}). Separating these readings requires a design that varies the diagnosis-presentation format and the student model systematically, which is beyond what a single null can settle.

These two findings belong together because the design orthogonalizes generation from scoring. Exp.~3 fixes the dialogues and varies only the scoring criterion, isolating criterion effects from tutor outputs; Exp.~4 fixes the evaluator and varies only the information supplied to the tutor, isolating pipeline-information effects from evaluator choice. The central comparison is therefore not a loose juxtaposition across experiments, but a paired audit of how much measured success moves when we change the scorer versus when we change the tutor's information.

\textbf{Implications}: Cross-experiment convergence anchors LLM-judge RSR as a property of the instrument---Exp.~3 (78--84\%) and Exp.~4 (74--78\%) settle near ${\sim}$79\% under independent pipeline conditions. Given that benchmark-style strict exact-match can substantially lower measured remediation estimates under this protocol, researchers should report RSR under at least two automated criteria, quantify their disagreement, and include a human baseline to anchor absolute values. When criteria disagree, the disagreement set should be audited directly rather than collapsed into a single preferred score. Without this discipline of multi-criterion scoring, multi-judge convergence, and human calibration, no single RSR figure can support a coherent interpretation.

\section{Conclusion}
\label{sec:conclusion}

MisconceptDialogue separates a \emph{capability} question from a \emph{measurement} question across five 7B--14B LLM tutors. One finding organizes the rest: on the same 2,250 dialogues, changing only the scoring criterion moves RSR from 42.1\% to 81.3\%, and that 39.2\,pp gap exceeds the combined effects of tutor model, teaching strategy, and supplied diagnostic information. What most determined measured success was therefore not any modelling lever, but the evaluator's definition of success. This is not the artifact of one lenient judge---four judges spanning three vendor families reproduce the gap, and 200 human annotations anchor it (judge--human $\kappa$=0.74--0.91). On capability, structured pipeline information yielded no detectable advantage over direct tutoring (GLMM $p$=0.263), and a wrong-diagnosis control found little evidence that tutors condition on the diagnostic labels they are handed. Reporting for tutoring evaluation should therefore treat the scoring criterion as a first-class object of study rather than an implementation detail. The calibrated instrument this protocol provides is what a real-student study will need first; that classroom study is the essential next step.

\phantomsection
\section*{Limitations}
\label{sec:limitations}

The limitations below primarily constrain external validity to real classrooms, broader domains, and larger model families; our central claims rest on within-protocol contrasts.

\begin{itemize}[leftmargin=*,itemsep=3pt,topsep=2pt,parsep=0pt]

\item \textbf{Domain and taxonomy.} We focus on mathematics for its relatively clear-cut answers; more ambiguous domains would require adapted criteria. The 55 educator-validated middle-school algebra misconceptions suffice for the type-level effects we report but are a small slice of real curricula.

\item \textbf{Dataset construction.} The detection datasets are imbalanced (BEA and Bridge contain only incorrect responses; MathDial's split is partial), a skew quantified in Exp.~1; relatedly, the Exp.~4 C4 detection rate (96.8\%) is recall, not specificity. The GPT-5.4-generated 47\% of remediation scenarios, combined with the OpenAI-family default judge, could inflate RSR via in-family coherence; the clean-set human RSR (88\%) and cross-vendor replication (Fleiss $\kappa$=0.85) bound but do not eliminate this risk.

\item \textbf{Model scale.} We evaluate five 7B--14B open-weight tutors plus two stronger non-OpenAI proprietary tutors (\S\ref{sec:exp5}); model identity contributes negligible variance (ICC$<$1\%). The out-of-family null suggests---but does not verify---that 70B+ systems behave similarly.

\item \textbf{Sampling, power, and human calibration.} Exp.~3 uses 30 dialogues per cell (2,250 total): per-cell power is moderate, though the paired criterion comparisons carrying our central finding remain highly significant. The Exp.~4 null is bounded rather than exact: TOST accepts equivalence at $\delta$=10\,pp for all 15 pairs but only three at $\delta$=5\,pp, and power is 80\% for effects $\geq$7\,pp, so benefits near 5\,pp remain possible. The human study covers two $n$=100 quality-stratified subsets; broader annotation and external annotator pools would tighten calibration (the clean-set human RSR of 88\% partly reflects quality screening).

\item \textbf{Judge--simulator dual role.} GPT-5-mini doubles as simulated student and as one of four LLM judges in Exp.~3--4. We probe this through cross-family validation (\S\ref{sec:exp4}), non-OpenAI judge replication (Fleiss $\kappa$=0.85), and human calibration on 200 dialogues (all four judges reach $\kappa\geq$0.74 against human consensus). Structural biases shared across the LLM-judge class are only partially anchored by this calibration.

\item \textbf{Diagnosis-presentation format.} The diagnosis is injected into tutor prompts in a single format throughout, so the wrong-diagnosis control tests conditioning under that format only and the claim is scoped accordingly (Appendix~\ref{sec:appendix-wrong-diag} reports a diagnosis-emphasis robustness check).

\item \textbf{Strategy invariance.} The 9--19\,pp within-model strategy gap admits two readings: genuine pedagogical equivalence at this scale, or a uniformly persuadable simulator. Restricting the contrast to the three faithful strategies (Appendix~\ref{sec:appendix-fidelity}) preserves the near-flat pattern; separating the two requires real students.

\end{itemize}

\section*{Ethical Considerations}

This work uses publicly available or research-use external datasets (including MathDial under CC BY-SA 4.0 and Bridge under CC BY 4.0) for research evaluation and involves no real student participants. The datasets contain mathematics problems and tutoring dialogues with no personally identifying information, so no offensive-content filtering or anonymization beyond the original releases was required; human annotation was judge-blinded and is reported only in aggregated, anonymized form (\S\ref{sec:human-eval}, Appendix~\ref{sec:appendix-humaneval-detail}). Simulating students sidesteps deployment risks with real learners, but it may overestimate real-classroom effectiveness, since simulated students cannot fully capture diverse learners' reasoning, language backgrounds, or affective responses. The more direct risk is \emph{inaccurate evaluation}: a lenient automated judge could certify a weak tutor as effective, and our 39.2\,pp criterion-induced shift shows how readily evaluation design can create a false sense of safety. We therefore recommend human expert review before autonomous deployment, particularly in high-stakes educational contexts.

\section*{Acknowledgements}
This work was supported by Institute of Information and Communications Technology Planning and Evaluation (IITP) grant funded by the Korea government (MSIT) (RS-2019-II190004, Development of semi-supervised learning language intelligence technology and Korean tutoring service for foreigners).

\bibliography{custom}

@article{schuirmann1987comparison,
  title={A comparison of the two one-sided tests procedure and the power approach for assessing the equivalence of average bioavailability},
  author={Schuirmann, Donald J},
  journal={Journal of pharmacokinetics and biopharmaceutics},
  volume={15},
  number={6},
  pages={657--680},
  year={1987},
  publisher={Springer}
}

@inproceedings{feldman2018automatic,
  title={Automatic Diagnosis of Students' Misconceptions in {K-8} Mathematics},
  author={Feldman, Molly Q and Cho, Ji Yong and Ong, Monica and Gulwani, Sumit and Popovi{\'c}, Zoran and Andersen, Erik},
  booktitle={Proceedings of the 2018 CHI Conference on Human Factors in Computing Systems},
  pages={1--12},
  year={2018}
}

@inproceedings{mcnichols2023algebra,
  title={Algebra Error Classification with Large Language Models},
  author={McNichols, Hunter and Zhang, Mengxue and Lan, Andrew},
  booktitle={International Conference on Artificial Intelligence in Education},
  pages={365--376},
  year={2023},
  organization={Springer}
}

@inproceedings{kochmar2025findings,
  title = "Findings of the {BEA} 2025 Shared Task on Pedagogical Ability Assessment of {AI}-powered Tutors",
    author = {Kochmar, Ekaterina  and
      Maurya, Kaushal Kumar  and
      Petukhova, Kseniia  and
      Srivatsa, KV Aditya  and
      Tack, Ana{\"i}s  and
      Vasselli, Justin},
    editor = {Kochmar, Ekaterina  and
      Alhafni, Bashar  and
      Bexte, Marie  and
      Burstein, Jill  and
      Horbach, Andrea  and
      Laarmann-Quante, Ronja  and
      Tack, Ana{\"i}s  and
      Yaneva, Victoria  and
      Yuan, Zheng},
    booktitle = "Proceedings of the 20th Workshop on Innovative Use of NLP for Building Educational Applications (BEA 2025)",
    month = jul,
    year = "2025",
    address = "Vienna, Austria",
    publisher = "Association for Computational Linguistics",
    url = "https://aclanthology.org/2025.bea-1.77/",
    doi = "10.18653/v1/2025.bea-1.77",
    pages = "1011--1033",
    ISBN = "979-8-89176-270-1"
}

@misc{king2024eedi,
  title={{EEDI -- Mining Misconceptions in Mathematics}},
  author={King, Jules and Burleigh, L and Woodhead, Simon and Kon, Panagiota and Baffour, Perpetual and Crossley, Scott and Reade, Walter and Demkin, Maggie},
  year={2024},
  howpublished={Kaggle competition},
  url={https://www.kaggle.com/competitions/eedi-mining-misconceptions-in-mathematics}
}

@article{otero2025benchmark,
  title={A Benchmark for Math Misconceptions: Bridging Gaps in Middle School Algebra with {AI}-Supported Instruction},
  author={Otero, Nancy and Druga, Stefania and Lan, Andrew},
  journal={Discover Education},
  volume={4},
  number={1},
  pages={277},
  year={2025},
  publisher={Springer}
}

@inproceedings{wang2021results,
  title={Results and Insights from Diagnostic Questions: The {NeurIPS} 2020 Education Challenge},
  author={Wang, Zichao and Lamb, Angus and Saveliev, Evgeny and Cameron, Pashmina and Zaykov, Jordan and Hernandez-Lobato, Jose Miguel and Turner, Richard E and Baraniuk, Richard G and Barton, Craig and Jones, Simon Peyton and Woodhead, Simon and Zhang, Cheng},
  booktitle={NeurIPS 2020 Competition and Demonstration Track},
  pages={191--205},
  year={2021},
  organization={PMLR}
}

@inproceedings{macina2025mathtutorbench,
  title={{MathTutorBench}: A Benchmark for Measuring Open-Ended Pedagogical Capabilities of {LLM} Tutors},
  author={Macina, Jakub and Daheim, Nico and Hakimi, Ido and Kapur, Manu and Gurevych, Iryna and Sachan, Mrinmaya},
  booktitle={Proceedings of the 2025 Conference on Empirical Methods in Natural Language Processing},
  pages={204--221},
  year={2025}
}

@article{liu2025discerning,
  title={Discerning Minds or Generic Tutors? Evaluating Instructional Guidance Capabilities in {Socratic} {LLMs}},
  author={Liu, Ying and Li, Can and Zhang, Ting and Wang, Mei and Zhu, Qiannan and Li, Jian and Huang, Hua},
  journal={arXiv preprint arXiv:2508.06583},
  year={2025}
}

@inproceedings{shi2026solver,
  title={From Solver to Tutor: Evaluating the Pedagogical Intelligence of {LLMs} with {KMP-Bench}},
  author={Shi, Weikang and Ren, Houxing and Pan, Junting and Zhou, Aojun and Wang, Ke and Lu, Zimu and Yang, Yunqiao and Hu, Yuxuan and Wei, Linda and Zhan, Mingjie and Li, Hongsheng},
  booktitle={Proceedings of the AAAI Conference on Artificial Intelligence},
  volume={40},
  number={39},
  pages={32965--32973},
  year={2026}
}

@inproceedings{hsu2026mathedu,
  title={{MathEDU}: Feedback Generation on Problem-Solving Processes for Mathematical Learning Support},
  author={Hsu, Wei-Ling and Tang, Yu-Chien and Yen, An-Zi},
  booktitle={Proceedings of the 19th Conference of the European Chapter of the Association for Computational Linguistics (Volume 1: Long Papers)},
  pages={2883--2901},
  year={2026}
}

@inproceedings{peng2025kele,
  title={{KELE}: A Multi-Agent Framework for Structured {Socratic} Teaching with Large Language Models},
  author={Peng, Xian and Yuan, Pan and Li, Dong and Cheng, Junlong and Fang, Qin and Liu, Zhi},
  booktitle={Findings of the Association for Computational Linguistics: EMNLP 2025},
  pages={16342--16362},
  year={2025}
}

@inproceedings{dinucu2025problem,
  title={From Problem-Solving to Teaching Problem-Solving: Aligning {LLMs} with Pedagogy Using Reinforcement Learning},
  author={Dinucu-Jianu, David and Macina, Jakub and Daheim, Nico and Hakimi, Ido and Gurevych, Iryna and Sachan, Mrinmaya},
  booktitle={Proceedings of the 2025 Conference on Empirical Methods in Natural Language Processing},
  pages={272--292},
  year={2025}
}

@article{team2024learnlm,
  title={{LearnLM}: Improving {Gemini} for Learning},
  author={Team, LearnLM and Modi, Abhinit and Veerubhotla, Aditya Srikanth and Rysbek, Aliya and Huber, Andrea and Wiltshire, Brett and Veprek, Brian and Gillick, Daniel and Kasenberg, Daniel and Ahmed, Derek and others},
  journal={arXiv preprint arXiv:2412.16429},
  year={2024}
}

@inproceedings{borchers2025can,
  title={Can Large Language Models Match Tutoring System Adaptivity? A Benchmarking Study},
  author={Borchers, Conrad and Shou, Tianze},
  booktitle={International Conference on Artificial Intelligence in Education},
  pages={407--420},
  year={2025},
  organization={Springer}
}

@inproceedings{macina2023mathdial,
  title={{MathDial}: A Dialogue Tutoring Dataset with Rich Pedagogical Properties Grounded in Math Reasoning Problems},
  author={Macina, Jakub and Daheim, Nico and Chowdhury, Sankalan and Sinha, Tanmay and Kapur, Manu and Gurevych, Iryna and Sachan, Mrinmaya},
  booktitle={Findings of the Association for Computational Linguistics: EMNLP 2023},
  pages={5602--5621},
  year={2023}
}

@inproceedings{wang2024bridging,
  title={Bridging the Novice-Expert Gap via Models of Decision-Making: A Case Study on Remediating Math Mistakes},
  author={Wang, Rose and Zhang, Qingyang and Robinson, Carly and Loeb, Susanna and Demszky, Dorottya},
  booktitle={Proceedings of the 2024 Conference of the North American Chapter of the Association for Computational Linguistics: Human Language Technologies (Volume 1: Long Papers)},
  pages={2174--2199},
  year={2024}
}

@article{liu2024socraticlm,
  title={{SocraticLM}: Exploring {Socratic} Personalized Teaching with Large Language Models},
  author={Liu, Jiayu and Huang, Zhenya and Xiao, Tong and Sha, Jing and Wu, Jinze and Liu, Qi and Wang, Shijin and Chen, Enhong},
  journal={Advances in Neural Information Processing Systems},
  volume={37},
  pages={85693--85721},
  year={2024}
}

@misc{sharma2026convolearn,
  title={{ConvoLearn}: A Learning Sciences Grounded Dataset for Fine-Tuning Dialogic {AI} Tutors}, 
      author={Mayank Sharma and Roy Pea and Hari Subramonyam},
      year={2026},
      eprint={2601.08950},
      archivePrefix={arXiv},
      primaryClass={cs.AI},
      url={https://arxiv.org/abs/2601.08950}, 
}

@article{chi2005commonsense,
  title={Commonsense conceptions of emergent processes: Why some misconceptions are robust},
  author={Chi, Michelene TH},
  journal={The journal of the learning sciences},
  volume={14},
  number={2},
  pages={161--199},
  year={2005},
  publisher={Taylor \& Francis}
}

@article{posner1982accommodation,
  title={Accommodation of a scientific conception: Toward a theory of conceptual change},
  author={Posner, George J and Strike, Kenneth A and Hewson, Peter W and Gertzog, William A},
  journal={Science education},
  volume={66},
  number={2},
  pages={211--227},
  year={1982},
  publisher={Hoboken}
}

@article{vosniadou2004extending,
  title={Extending the conceptual change approach to mathematics learning and teaching.},
  author={Vosniadou, Stella and Verschaffel, Lieven},
  journal={Learning and Instruction},
  volume={14},
  number={5},
  pages={445--451},
  year={2004},
  publisher={Elsevier}
}

@article{smith1994misconceptions,
  title={Misconceptions reconceived: A constructivist analysis of knowledge in transition},
  author={Smith III, John P and DiSessa, Andrea A and Roschelle, Jeremy},
  journal={The journal of the learning sciences},
  volume={3},
  number={2},
  pages={115--163},
  year={1994},
  publisher={Taylor \& Francis}
}

@article{tippett2010refutation,
  title={Refutation text in science education: A review of two decades of research},
  author={Tippett, Christine D},
  journal={International journal of science and mathematics education},
  volume={8},
  number={6},
  pages={951--970},
  year={2010},
  publisher={Springer}
}

@article{confrey1990chapter,
  title={Chapter 1: A review of the research on student conceptions in mathematics, science, and programming},
  author={Confrey, Jere},
  journal={Review of research in education},
  volume={16},
  number={1},
  pages={3--56},
  year={1990},
  publisher={Sage Publications Sage CA: Thousand Oaks, CA}
}

@article{vanlehn2011relative,
  title={The relative effectiveness of human tutoring, intelligent tutoring systems, and other tutoring systems},
  author={VanLehn, Kurt},
  journal={Educational psychologist},
  volume={46},
  number={4},
  pages={197--221},
  year={2011},
  publisher={Taylor \& Francis}
}

@article{kulik2016effectiveness,
  title={Effectiveness of intelligent tutoring systems: a meta-analytic review},
  author={Kulik, James A and Fletcher, John D},
  journal={Review of educational research},
  volume={86},
  number={1},
  pages={42--78},
  year={2016},
  publisher={Sage Publications Sage CA: Los Angeles, CA}
}

@article{nye2014autotutor,
  title={{AutoTutor} and Family: A Review of 17 Years of Natural Language Tutoring},
  author={Nye, Benjamin D and Graesser, Arthur C and Hu, Xiangen},
  journal={International Journal of Artificial Intelligence in Education},
  volume={24},
  number={4},
  pages={427--469},
  year={2014},
  publisher={Springer}
}

@inproceedings{markel2023gpteach,
  title={{GPTeach}: Interactive {TA} Training with {GPT}-Based Students},
  author={Markel, Julia M and Opferman, Steven G and Landay, James A and Piech, Chris},
  booktitle={Proceedings of the tenth acm conference on learning@ scale},
  pages={226--236},
  year={2023}
}

@article{graesser2004autotutor,
  title={{AutoTutor}: A Tutor with Dialogue in Natural Language},
  author={Graesser, Arthur C and Lu, Shulan and Jackson, George Tanner and Mitchell, Heather Hite and Ventura, Mathew and Olney, Andrew and Louwerse, Max M},
  journal={Behavior Research Methods, Instruments, \& Computers},
  volume={36},
  number={2},
  pages={180--192},
  year={2004},
  publisher={Springer}
}

@misc{qwen2024qwen25technicalreport,
      title={{Qwen2.5} Technical Report}, 
      author={An Yang and Baosong Yang and Beichen Zhang and Binyuan Hui and Bo Zheng and Bowen Yu and Chengyuan Li and Dayiheng Liu and Fei Huang and Haoran Wei and Huan Lin and Jian Yang and Jianhong Tu and Jianwei Zhang and Jianxin Yang and Jiaxi Yang and Jingren Zhou and Junyang Lin and Kai Dang and Keming Lu and Keqin Bao and Kexin Yang and Le Yu and Mei Li and Mingfeng Xue and Pei Zhang and Qin Zhu and Rui Men and Runji Lin and Tianhao Li and Tianyi Tang and Tingyu Xia and Xingzhang Ren and Xuancheng Ren and Yang Fan and Yang Su and Yichang Zhang and Yu Wan and Yuqiong Liu and Zeyu Cui and Zhenru Zhang and Zihan Qiu},
      year={2024},
      eprint={2412.15115},
      archivePrefix={arXiv},
      primaryClass={cs.CL},
      url={https://arxiv.org/abs/2412.15115}, 
}

@misc{singh2025openai,
  title={{OpenAI} {GPT-5} System Card}, 
      author={Aaditya Singh and Adam Fry and Adam Perelman and Adam Tart and Adi Ganesh and Ahmed El-Kishky and Aidan McLaughlin and Aiden Low and AJ Ostrow and Akhila Ananthram and Akshay Nathan and Alan Luo and Alec Helyar and Aleksander Madry and Aleksandr Efremov and Aleksandra Spyra and Alex Baker-Whitcomb and Alex Beutel and Alex Karpenko and Alex Makelov and Alex Neitz and Alex Wei and Alexandra Barr and Alexandre Kirchmeyer and Alexey Ivanov and Alexi Christakis and Alistair Gillespie and Allison Tam and Ally Bennett and Alvin Wan and Alyssa Huang and Amy McDonald Sandjideh and Amy Yang and Ananya Kumar and Andre Saraiva and Andrea Vallone and Andrei Gheorghe and Andres Garcia Garcia and Andrew Braunstein and Andrew Liu and Andrew Schmidt and Andrey Mereskin and Andrey Mishchenko and Andy Applebaum and Andy Rogerson and Ann Rajan and Annie Wei and Anoop Kotha and Anubha Srivastava and Anushree Agrawal and Arun Vijayvergiya and Ashley Tyra and Ashvin Nair and Avi Nayak and Ben Eggers and Bessie Ji and Beth Hoover and Bill Chen and Blair Chen and Boaz Barak and Borys Minaiev and Botao Hao and Bowen Baker and Brad Lightcap and Brandon McKinzie and Brandon Wang and Brendan Quinn and Brian Fioca and Brian Hsu and Brian Yang and Brian Yu and Brian Zhang and Brittany Brenner and Callie Riggins Zetino and Cameron Raymond and Camillo Lugaresi and Carolina Paz and Cary Hudson and Cedric Whitney and Chak Li and Charles Chen and Charlotte Cole and Chelsea Voss and Chen Ding and Chen Shen and Chengdu Huang and Chris Colby and Chris Hallacy and Chris Koch and Chris Lu and Christina Kaplan and Christina Kim and CJ Minott-Henriques and Cliff Frey and Cody Yu and Coley Czarnecki and Colin Reid and Colin Wei and Cory Decareaux and Cristina Scheau and Cyril Zhang and Cyrus Forbes and Da Tang and Dakota Goldberg and Dan Roberts and Dana Palmie and Daniel Kappler and Daniel Levine and Daniel Wright and Dave Leo and David Lin and David Robinson and Declan Grabb and Derek Chen and Derek Lim and Derek Salama and Dibya Bhattacharjee and Dimitris Tsipras and Dinghua Li and Dingli Yu and DJ Strouse and Drew Williams and Dylan Hunn and Ed Bayes and Edwin Arbus and Ekin Akyurek and Elaine Ya Le and Elana Widmann and Eli Yani and Elizabeth Proehl and Enis Sert and Enoch Cheung and Eri Schwartz and Eric Han and Eric Jiang and Eric Mitchell and Eric Sigler and Eric Wallace and Erik Ritter and Erin Kavanaugh and Evan Mays and Evgenii Nikishin and Fangyuan Li and Felipe Petroski Such and Filipe de Avila Belbute Peres and Filippo Raso and Florent Bekerman and Foivos Tsimpourlas and Fotis Chantzis and Francis Song and Francis Zhang and Gaby Raila and Garrett McGrath and Gary Briggs and Gary Yang and Giambattista Parascandolo and Gildas Chabot and Grace Kim and Grace Zhao and Gregory Valiant and Guillaume Leclerc and Hadi Salman and Hanson Wang and Hao Sheng and Haoming Jiang and Haoyu Wang and Haozhun Jin and Harshit Sikchi and Heather Schmidt and Henry Aspegren and Honglin Chen and Huida Qiu and Hunter Lightman and Ian Covert and Ian Kivlichan and Ian Silber and Ian Sohl and Ibrahim Hammoud and Ignasi Clavera and Ikai Lan and Ilge Akkaya and Ilya Kostrikov and Irina Kofman and Isak Etinger and Ishaan Singal and Jackie Hehir and Jacob Huh and Jacqueline Pan and Jake Wilczynski and Jakub Pachocki and James Lee and James Quinn and Jamie Kiros and Janvi Kalra and Jasmyn Samaroo and Jason Wang and Jason Wolfe and Jay Chen and Jay Wang and Jean Harb and Jeffrey Han and Jeffrey Wang and Jennifer Zhao and Jeremy Chen and Jerene Yang and Jerry Tworek and Jesse Chand and Jessica Landon and Jessica Liang and Ji Lin and Jiancheng Liu and Jianfeng Wang and Jie Tang and Jihan Yin and Joanne Jang and Joel Morris and Joey Flynn and Johannes Ferstad and Johannes Heidecke and John Fishbein and John Hallman and Jonah Grant and Jonathan Chien and Jonathan Gordon and Jongsoo Park and Jordan Liss and Jos Kraaijeveld and Joseph Guay and Joseph Mo and Josh Lawson and Josh McGrath and Joshua Vendrow and Joy Jiao and Julian Lee and Julie Steele and Julie Wang and Junhua Mao and Kai Chen and Kai Hayashi and Kai Xiao and Kamyar Salahi and Kan Wu and Karan Sekhri and Karan Sharma and Karan Singhal and Karen Li and Kenny Nguyen and Keren Gu-Lemberg and Kevin King and Kevin Liu and Kevin Stone and Kevin Yu and Kristen Ying and Kristian Georgiev and Kristie Lim and Kushal Tirumala and Kyle Miller and Lama Ahmad and Larry Lv and Laura Clare and Laurance Fauconnet and Lauren Itow and Lauren Yang and Laurentia Romaniuk and Leah Anise and Lee Byron and Leher Pathak and Leon Maksin and Leyan Lo and Leyton Ho and Li Jing and Liang Wu and Liang Xiong and Lien Mamitsuka and Lin Yang and Lindsay McCallum and Lindsey Held and Liz Bourgeois and Logan Engstrom and Lorenz Kuhn and Louis Feuvrier and Lu Zhang and Lucas Switzer and Lukas Kondraciuk and Lukasz Kaiser and Manas Joglekar and Mandeep Singh and Mandip Shah and Manuka Stratta and Marcus Williams and Mark Chen and Mark Sun and Marselus Cayton and Martin Li and Marvin Zhang and Marwan Aljubeh and Matt Nichols and Matthew Haines and Max Schwarzer and Mayank Gupta and Meghan Shah and Melody Y. Guan and Melody Huang and Meng Dong and Mengqing Wang and Mia Glaese and Micah Carroll and Michael Lampe and Michael Malek and Michael Sharman and Michael Zhang and Michele Wang and Michelle Pokrass and Mihai Florian and Mikhail Pavlov and Miles Wang and Ming Chen and Mingxuan Wang and Minnia Feng and Mo Bavarian and Molly Lin and Moose Abdool and Mostafa Rohaninejad and Nacho Soto and Natalie Staudacher and Natan LaFontaine and Nathan Marwell and Nelson Liu and Nick Preston and Nick Turley and Nicklas Ansman and Nicole Blades and Nikil Pancha and Nikita Mikhaylin and Niko Felix and Nikunj Handa and Nishant Rai and Nitish Keskar and Noam Brown and Ofir Nachum and Oleg Boiko and Oleg Murk and Olivia Watkins and Oona Gleeson and Pamela Mishkin and Patryk Lesiewicz and Paul Baltescu and Pavel Belov and Peter Zhokhov and Philip Pronin and Phillip Guo and Phoebe Thacker and Qi Liu and Qiming Yuan and Qinghua Liu and Rachel Dias and Rachel Puckett and Rahul Arora and Ravi Teja Mullapudi and Raz Gaon and Reah Miyara and Rennie Song and Rishabh Aggarwal and RJ Marsan and Robel Yemiru and Robert Xiong and Rohan Kshirsagar and Rohan Nuttall and Roman Tsiupa and Ronen Eldan and Rose Wang and Roshan James and Roy Ziv and Rui Shu and Ruslan Nigmatullin and Saachi Jain and Saam Talaie and Sam Altman and Sam Arnesen and Sam Toizer and Sam Toyer and Samuel Miserendino and Sandhini Agarwal and Sarah Yoo and Savannah Heon and Scott Ethersmith and Sean Grove and Sean Taylor and Sebastien Bubeck and Sever Banesiu and Shaokyi Amdo and Shengjia Zhao and Sherwin Wu and Shibani Santurkar and Shiyu Zhao and Shraman Ray Chaudhuri and Shreyas Krishnaswamy and Shuaiqi and Xia and Shuyang Cheng and Shyamal Anadkat and Simón Posada Fishman and Simon Tobin and Siyuan Fu and Somay Jain and Song Mei and Sonya Egoian and Spencer Kim and Spug Golden and SQ Mah and Steph Lin and Stephen Imm and Steve Sharpe and Steve Yadlowsky and Sulman Choudhry and Sungwon Eum and Suvansh Sanjeev and Tabarak Khan and Tal Stramer and Tao Wang and Tao Xin and Tarun Gogineni and Taya Christianson and Ted Sanders and Tejal Patwardhan and Thomas Degry and Thomas Shadwell and Tianfu Fu and Tianshi Gao and Timur Garipov and Tina Sriskandarajah and Toki Sherbakov and Tomek Korbak and Tomer Kaftan and Tomo Hiratsuka and Tongzhou Wang and Tony Song and Tony Zhao and Troy Peterson and Val Kharitonov and Victoria Chernova and Vineet Kosaraju and Vishal Kuo and Vitchyr Pong and Vivek Verma and Vlad Petrov and Wanning Jiang and Weixing Zhang and Wenda Zhou and Wenlei Xie and Wenting Zhan and Wes McCabe and Will DePue and Will Ellsworth and Wulfie Bain and Wyatt Thompson and Xiangning Chen and Xiangyu Qi and Xin Xiang and Xinwei Shi and Yann Dubois and Yaodong Yu and Yara Khakbaz and Yifan Wu and Yilei Qian and Yin Tat Lee and Yinbo Chen and Yizhen Zhang and Yizhong Xiong and Yonglong Tian and Young Cha and Yu Bai and Yu Yang and Yuan Yuan and Yuanzhi Li and Yufeng Zhang and Yuguang Yang and Yujia Jin and Yun Jiang and Yunyun Wang and Yushi Wang and Yutian Liu and Zach Stubenvoll and Zehao Dou and Zheng Wu and Zhigang Wang},
      year={2026},
      eprint={2601.03267},
      archivePrefix={arXiv},
      primaryClass={cs.CL},
      url={https://arxiv.org/abs/2601.03267}, 
}

@article{grattafiori2024llama,
  title={The {Llama} 3 Herd of Models},
  author={Grattafiori, Aaron and Dubey, Abhimanyu and Jauhri, Abhinav and Pandey, Abhinav and Kadian, Abhishek and Al-Dahle, Ahmad and Letman, Aiesha and Mathur, Akhil and Schelten, Alan and Vaughan, Alex and others},
  journal={arXiv preprint arXiv:2407.21783},
  year={2024}
}

@misc{jiang2023mistral7b,
      title={{Mistral 7B}}, 
      author={Albert Q. Jiang and Alexandre Sablayrolles and Arthur Mensch and Chris Bamford and Devendra Singh Chaplot and Diego de las Casas and Florian Bressand and Gianna Lengyel and Guillaume Lample and Lucile Saulnier and Lélio Renard Lavaud and Marie-Anne Lachaux and Pierre Stock and Teven Le Scao and Thibaut Lavril and Thomas Wang and Timothée Lacroix and William El Sayed},
      year={2023},
      eprint={2310.06825},
      archivePrefix={arXiv},
      primaryClass={cs.CL},
      url={https://arxiv.org/abs/2310.06825}, 
}

@article{abdin2024phi,
  title={{Phi-4} Technical Report},
  author={Abdin, Marah and Aneja, Jyoti and Behl, Harkirat and Bubeck, S{\'e}bastien and Eldan, Ronen and Gunasekar, Suriya and Harrison, Michael and Hewett, Russell J and Javaheripi, Mojan and Kauffmann, Piero and others},
  journal={arXiv preprint arXiv:2412.08905},
  year={2024}
}

@misc{yang2025qwen3technicalreport,
      title={{Qwen3} Technical Report}, 
      author={An Yang and Anfeng Li and Baosong Yang and Beichen Zhang and Binyuan Hui and Bo Zheng and Bowen Yu and Chang Gao and Chengen Huang and Chenxu Lv and Chujie Zheng and Dayiheng Liu and Fan Zhou and Fei Huang and Feng Hu and Hao Ge and Haoran Wei and Huan Lin and Jialong Tang and Jian Yang and Jianhong Tu and Jianwei Zhang and Jianxin Yang and Jiaxi Yang and Jing Zhou and Jingren Zhou and Junyang Lin and Kai Dang and Keqin Bao and Kexin Yang and Le Yu and Lianghao Deng and Mei Li and Mingfeng Xue and Mingze Li and Pei Zhang and Peng Wang and Qin Zhu and Rui Men and Ruize Gao and Shixuan Liu and Shuang Luo and Tianhao Li and Tianyi Tang and Wenbiao Yin and Xingzhang Ren and Xinyu Wang and Xinyu Zhang and Xuancheng Ren and Yang Fan and Yang Su and Yichang Zhang and Yinger Zhang and Yu Wan and Yuqiong Liu and Zekun Wang and Zeyu Cui and Zhenru Zhang and Zhipeng Zhou and Zihan Qiu},
      year={2025},
      eprint={2505.09388},
      archivePrefix={arXiv},
      primaryClass={cs.CL},
      url={https://arxiv.org/abs/2505.09388}, 
}

@inproceedings{xiao2024c,
  title={{C-Pack}: Packed Resources for General {Chinese} Embeddings},
  author={Xiao, Shitao and Liu, Zheng and Zhang, Peitian and Muennighoff, Niklas and Lian, Defu and Nie, Jian-Yun},
  booktitle={Proceedings of the 47th international ACM SIGIR conference on research and development in information retrieval},
  pages={641--649},
  year={2024}
}

@article{lewis2020retrieval,
  title={Retrieval-Augmented Generation for Knowledge-Intensive {NLP} Tasks},
  author={Lewis, Patrick and Perez, Ethan and Piktus, Aleksandra and Petroni, Fabio and Karpukhin, Vladimir and Goyal, Naman and K{\"u}ttler, Heinrich and Lewis, Mike and Yih, Wen-tau and Rockt{\"a}schel, Tim and Riedel, Sebastian and Kiela, Douwe},
  journal={Advances in neural information processing systems},
  volume={33},
  pages={9459--9474},
  year={2020}
}

@article{wei2022cot,
    title = {Chain-of-Thought Prompting Elicits Reasoning in Large Language Models},
    author = {Wei, Jason and Wang, Xuezhi and Schuurmans, Dale and Bosma, Maarten and Ichter, Brian and Xia, Fei and Chi, Ed and Le, Quoc and Zhou, Denny},
    journal = {Advances in Neural Information Processing Systems},
    volume = {35},
    pages = {24824--24837},
    year = {2022},
}

@article{zheng2023judging,
  title={Judging {LLM}-as-a-Judge with {MT-Bench} and Chatbot Arena},
  author={Zheng, Lianmin and Chiang, Wei-Lin and Sheng, Ying and Zhuang, Siyuan and Wu, Zhanghao and Zhuang, Yonghao and Lin, Zi and Li, Zhuohan and Li, Dacheng and Xing, Eric P and Zhang, Hao and Gonzalez, Joseph E and Stoica, Ion},
  journal={Advances in neural information processing systems},
  volume={36},
  pages={46595--46623},
  year={2023}
}

@inproceedings{mitton2026genretrieve,
    title = {Misconception Diagnosis From Student-Tutor Dialogue: Generate, Retrieve, Rerank},
    author = {Mitton, Joshua and Bhattacharyya, Prarthana and Smith, Digory and Christie, Thomas and Abboud, Ralph and Woodhead, Simon},
    booktitle = {Proceedings of the Thirteenth ACM Conference on Learning @ Scale},
    pages = {277--286},
    year = {2026},
    publisher = {ACM},
}

@article{panickssery2024llm,
  title={{LLM} Evaluators Recognize and Favor Their Own Generations},
  author={Panickssery, Arjun and Bowman, Samuel R and Feng, Shi},
  journal={Advances in Neural Information Processing Systems},
  volume={37},
  pages={68772--68802},
  year={2024}
}

@article{kasneci2023chatgpt,
  title={{ChatGPT} for Good? On Opportunities and Challenges of Large Language Models for Education},
  author={Kasneci, Enkelejda and Se{\ss}ler, Kathrin and K{\"u}chemann, Stefan and Bannert, Maria and Dementieva, Daryna and Fischer, Frank and Gasser, Urs and Groh, Georg and G{\"u}nnemann, Stephan and H{\"u}llermeier, Eyke and Krusche, Stephan and Kutyniok, Gitta and Michaeli, Tilman and Nerdel, Claudia and Pfeffer, J{\"u}rgen and Poquet, Oleksandra and Sailer, Michael and Schmidt, Albrecht and Seidel, Tina and Stadler, Matthias and Weller, Jochen and Kuhn, Jochen and Kasneci, Gjergji},
  journal={Learning and individual differences},
  volume={103},
  pages={102274},
  year={2023},
  publisher={Elsevier}
}

@inproceedings{tack2022ai,
  title={The {AI} Teacher Test: Measuring the Pedagogical Ability of {Blender} and {GPT-3} in Educational Dialogues},
  author={Tack, Ana{\"\i}s and Piech, Chris},
  booktitle={Proceedings of the 15th International Conference on Educational Data Mining},
  pages={522--529},
  year={2022}
}

@inproceedings{scarlatos2026simulated,
  title={Simulated Students in Tutoring Dialogues: Substance or Illusion?},
  author={Scarlatos, Alexander and Lee, Jaewook and Woodhead, Simon and Lan, Andrew},
  booktitle={Proceedings of the 64th Annual Meeting of the Association for Computational Linguistics (Volume 1: Long Papers)},
  pages={42349--42385},
  year={2026},
  publisher={Association for Computational Linguistics}
}

@inproceedings{vanzo2025gpt,
  title={{GPT-4} as a Homework Tutor Can Improve Student Engagement and Learning Outcomes},
  author={Vanzo, Alessandro and Chowdhury, Sankalan Pal and Sachan, Mrinmaya},
  booktitle={Proceedings of the 63rd Annual Meeting of the Association for Computational Linguistics (Volume 1: Long Papers)},
  pages={31119--31136},
  year={2025}
}

@inproceedings{wu2025embracing,
  title={Embracing Imperfection: Simulating Students with Diverse Cognitive Levels Using {LLM}-Based Agents},
  author={Wu, Tao and Chen, Jingyuan and Lin, Wang and Li, Mengze and Zhu, Yumeng and Li, Ang and Kuang, Kun and Wu, Fei},
  booktitle={Proceedings of the 63rd Annual Meeting of the Association for Computational Linguistics (Volume 1: Long Papers)},
  pages={9887--9908},
  year={2025}
}

@inproceedings{puech2025towards,
  title={Towards the Pedagogical Steering of Large Language Models for Tutoring: A Case Study with Modeling Productive Failure},
  author={Puech, Romain and Macina, Jakub and Chatain, Julia and Sachan, Mrinmaya and Kapur, Manu},
  booktitle={Findings of the Association for Computational Linguistics: ACL 2025},
  pages={26291--26311},
  year={2025}
}

@inproceedings{martynova2025can,
  title={Can {LLMs} Effectively Simulate Human Learners? Teachers' Insights from Tutoring {LLM} Students},
  author={Martynova, Daria and Macina, Jakub and Daheim, Nico and Yalcin, Nilay and Zhang, Xiaoyu and Sachan, Mrinmaya},
  booktitle={Proceedings of the 20th Workshop on Innovative Use of NLP for Building Educational Applications (BEA 2025)},
  pages={100--117},
  year={2025}
}

@inproceedings{chen2024humans,
  title={Humans or {LLMs} as the Judge? A Study on Judgement Bias},
  author={Chen, Guiming Hardy and Chen, Shunian and Liu, Ziche and Jiang, Feng and Wang, Benyou},
  booktitle={Proceedings of the 2024 Conference on Empirical Methods in Natural Language Processing},
  pages={8301--8327},
  year={2024}
}

@misc{gdm2026gemini,
  title = {{Gemini 3.1 Flash Lite Model Card}},
  author = {{Google DeepMind}},
  year = {2026},
  month = mar,
  howpublished = {Model card},
  url = {https://deepmind.google/models/model-cards/gemini-3-1-flash-lite/},
  note = {Released 2026-03-03}
}

@misc{anthropic2025haiku,
  title = {{Claude Haiku 4.5 System Card}},
  author = {{Anthropic}},
  year = {2025},
  month = oct,
  howpublished = {System card},
  url = {https://www.anthropic.com/claude-haiku-4-5-system-card},
  note = {Released 2025-10-15}
}

@article{kaser2024simulated,
  title={Simulated Learners in Educational Technology: A Systematic Literature Review and a {T}uring-like Test},
  author={K{\"a}ser, Tanja and Alexandron, Giora},
  journal={International Journal of Artificial Intelligence in Education},
  volume={34},
  number={2},
  pages={545--585},
  year={2024},
  publisher={Springer}
}

@article{heyueya2024psychometric,
  title={Psychometric Alignment: Capturing Human Knowledge Distributions via Language Models},
  author={He-Yueya, Joy and Ma, Wanjing Anya and Gandhi, Kanishk and Domingue, Benjamin W and Brunskill, Emma and Goodman, Noah D},
  journal={arXiv preprint arXiv:2407.15645},
  year={2024}
}

\appendix

\section{Prompt Specifications}
\label{sec:appendix-prompts}

We summarize the key prompt specifications used across the pipeline tasks and evaluation harness: detection, diagnosis, remediation, LLM-as-judge evaluation, simulated-student behavior, and quality screening.

\paragraph{Detection (Mistake identification).}
The prompt provides the problem statement and student response, then asks the model to classify the response as \texttt{correct}, \texttt{incorrect}, or \texttt{ambiguous}, with one-sentence reasoning.

\paragraph{Diagnosis (Misconception ranking).}
The prompt presents the student's incorrect response alongside the full misconception taxonomy, and asks the model to produce a ranked list of the most likely misconceptions with reasoning.

\paragraph{Remediation (Strategy-constrained).}
The prompt specifies the diagnosed misconception and the required strategy, then instructs the model to generate a tutoring response following that strategy without revealing the answer.

\paragraph{LLM-as-judge.} The system prompt instructs the model to assess whether the student demonstrated \emph{correct understanding} after remediation. Key distinction: ``correct understanding'' means conceptual grasp (true even with arithmetic slips), while ``correct answer'' means numerical correctness (false if reasoning is wrong despite correct answer). Output: JSON with \texttt{correct} (bool), \texttt{reasoning} (str), \texttt{confidence} (float), \texttt{misconception\_resolved} (bool). The user prompt provides the verification problem, correct answer, misconception, and student response. Judge models: GPT-5-mini (Exp.~3--4 default), GPT-5.4, Claude Haiku 4.5, and Gemini 3.1 Flash Lite (\S\ref{sec:exp3}).

\paragraph{Simulated student.} GPT-5-mini acts as the simulated student. The system prompt specifies (1)~knowledge level (low/medium/high), (2)~cooperation level (cooperative/resistant), (3)~the misconception to exhibit, and (4)~the student's incorrect answer. The student must behave consistently with the misconception but may gradually show understanding if the tutor is effective. After dialogue, the student responds to a verification problem at temperature=0.3.

\paragraph{Quality screening.} GPT-5.4 rates dialogue quality on a 1--5 scale across multiple dimensions: dialogue coherence, role adherence (each agent stays in its assigned role), educational value, and verification response relevance. Dialogues scoring $\geq$3.0 are eligible for the clean human evaluation set.

\section{Simulated Student Profiles}
\label{sec:appendix-profiles}

We use six simulated student profiles crossing three knowledge levels (low, medium, high) with two cooperation levels (cooperative, resistant), following \citet{scarlatos2026simulated}. Each profile specifies the student's knowledge level, behavioral tendencies, and the misconception they hold.

\section{Dataset Statistics}
\label{sec:appendix-data}

Table~\ref{tab:data-stats} summarizes the external datasets used across the four experiments. The separate Exp.~3/4 scenario bank contains 106 instantiated scenarios: 36 factual, 34 procedural, and 36 conceptual.

\begin{table}[h]
\centering
\small
\setlength{\tabcolsep}{4pt}
\begin{tabular}{@{}>{\raggedright\arraybackslash}p{0.40\columnwidth}rrl@{}}
\toprule
\textbf{Dataset} & \textbf{Instances} & \textbf{Turns} & \textbf{Source} \\
\midrule
BEA 2025 (dev) \cite{kochmar2025findings} & 300 & 4.2 & Shared Task \\
MathDial (train) \cite{macina2023mathdial} & 2{,}262 & 5.8 & GSM8K \\
Bridge (train) \cite{wang2024bridging} & 419 & 6.1 & Expert \\
Math Misconceptions Bench. \cite{otero2025benchmark} & 220 & n/a & Educator \\
\bottomrule
\end{tabular}
\caption{Dataset statistics. All datasets are external; Exp.~3/4 scenarios are categorized by the Math Misconceptions Benchmark's 55-misconception taxonomy.}
\label{tab:data-stats}
\end{table}

\section{Diagnosis Accuracy by Misconception Type}
\label{sec:appendix-bytype}

Table~\ref{tab:exp2-bytype} breaks down Exp.~2 diagnosis Acc@1 by misconception type across all five models and prompting conditions.

\begin{table*}[h]
\centering
\small
\adjustbox{max width=\textwidth}{%
\begin{tabular}{llccccc}
\toprule
\textbf{Condition} & \textbf{Type} & \textbf{Qwen-2.5} & \textbf{Mistral} & \textbf{Llama} & \textbf{Phi-4} & \textbf{Qwen3.5} \\
\midrule
\multirow{3}{*}{(A) Zero-shot}
& Factual    & 0.425 & 0.000 & 0.350 & 0.500 & \textbf{0.825} \\
& Procedural & 0.875 & 0.125 & 0.875 & 0.875 & \textbf{1.000} \\
& Conceptual & 0.320 & 0.093 & 0.250 & 0.401 & \textbf{0.459} \\
\midrule
\multirow{3}{*}{(C) CoT+rubric}
& Factual    & 0.575 & 0.200 & 0.325 & 0.500 & \textbf{0.650} \\
& Procedural & 0.750 & 0.500 & 0.625 & \textbf{0.875} & \textbf{0.875} \\
& Conceptual & 0.285 & 0.203 & 0.262 & 0.384 & \textbf{0.459} \\
\midrule
\multirow{3}{*}{(E) Gen-Ret-Rerank}
& Factual    & 0.525 & 0.250 & 0.350 & 0.575 & \textbf{0.600} \\
& Procedural & 0.750 & 0.375 & 0.750 & 0.750 & \textbf{0.875} \\
& Conceptual & 0.326 & 0.250 & 0.308 & 0.442 & \textbf{0.483} \\
\bottomrule
\end{tabular}%
}
\caption{Diagnosis Acc@1 by misconception type across five models after the uniform post-hoc parsing correction. Procedural misconceptions are easier to diagnose than conceptual ones for most models, though Exp.~2 type counts are imbalanced and the type-wise analysis is descriptive. Mistral remains the weakest diagnosis model but improves under retrieval-style conditions, while Qwen3.5 remains strongest overall and retains high procedural accuracy under both zero-shot and Gen-Ret-Rerank.}
\label{tab:exp2-bytype}
\end{table*}

\section{Dataset Label Distributions}
\label{sec:appendix-labels}

An important consideration for interpreting our Exp.~1 results: BEA 2025 and Bridge contain \emph{only} incorrect student responses, while MathDial has a less skewed distribution containing both correct and incorrect responses (1,696 correct, 566 incorrect; 75\%/25\% split). This imbalance explains the divergent context effects observed in Table~\ref{tab:exp1}: on all-incorrect datasets, multi-turn context helps the model learn to classify everything as incorrect, inflating both accuracy and F1.

\section{Experimental Hyperparameters}
\label{sec:appendix-hyperparams}

Table~\ref{tab:hyperparams} lists the key hyperparameters for each experimental stage. Local models are served via HuggingFace Transformers; GPT-5-mini is accessed through the OpenAI API.

\begin{table}[h]
\centering
\small
\adjustbox{max width=\columnwidth}{%
\begin{tabular}{lll}
\toprule
\textbf{Parameter} & \textbf{Value} & \textbf{Scope} \\
\midrule
Temperature (tutor) & 0.2 & Exp.~3, 4 \\
Temperature (student) & 0.7 & Exp.~3, 4 \\
Temperature (verifier) & 0.3 & Exp.~3, 4 \\
Max tokens (tutor) & 384 & Exp.~3, 4 \\
Max tokens (student) & 256 & Exp.~3, 4 \\
Max tokens (detection) & 2,048 & Exp.~1 \\
Top-$p$ & 0.95 & All \\
Dialogue turns & 3--5 & Exp.~3 \\
Dialogue turns & 5 max & Exp.~4 \\
Scenarios per cell & 30 & Exp.~3 \\
Total scenarios & 106 & Exp.~3, 4 \\
Dialogues & $5\times 5\times 3\times 30=2{,}250$ & Exp.~3 \\
Dialogues & $5\times 6\times 106=3{,}180$ & Exp.~4 \\
Random seed & 42 & All \\
\bottomrule
\end{tabular}%
}
\caption{Hyperparameters across experiments.}
\label{tab:hyperparams}
\end{table}

\section{Diagnosis Experiment Methodology}
\label{sec:appendix-diagnosis}

We detail the methodology for Exp.~2's five diagnosis conditions: data split, retrieval setup, the GRR pipeline, and post-hoc parsing correction.

\paragraph{Data split.} All 220 examples (55 misconceptions $\times$ 4 each) serve as the evaluation set with no separate train/test partition. Models perform a 55-class ranking task over the full taxonomy for each example.

\paragraph{RAG.} We encode the 55 taxonomy entries with BAAI/bge-base-en-v1.5 \cite{xiao2024c} (SentenceTransformer; batch size 64, L2-normalized). The query is the problem statement concatenated with the student's answer; we retrieve top-25 by cosine similarity and present them to the LLM with similarity scores for re-ranking. No separate reranking model is used.

\paragraph{Generate--Retrieve--Rerank (GRR).} A three-stage pipeline: (1)~\textit{Generate}: produce a free-text diagnosis of the student's error without consulting the taxonomy (max 256 tokens). (2)~\textit{Retrieve}: use the generated diagnosis as the retrieval query over the same bge-base-en-v1.5 index (top-25). (3)~\textit{Rerank}: present both the generated diagnosis and retrieved candidates to the LLM for final ranking (max 512 tokens). Rationale: the generated diagnosis text serves as a semantically richer query than raw problem--answer concatenation.

\paragraph{Post-hoc parsing correction.} Several models output misconception IDs in decorated form (e.g., ``1.\ MaE02 (conceptual, fractions) --- The student may\ldots''), depressing Acc@1 under strict string matching. We applied a uniform regex \texttt{Ma[A-Z]\textbackslash d\{2,3\}} to extract bare IDs; the correction was developed after inspecting results, but the rule is model-agnostic (matches only ID format) and applied uniformly to all models. Table~\ref{tab:parsing-correction} reports the per-condition before/after Acc@1.

\paragraph{Few-shot selection and leakage.} Few-shot examples are drawn from the same 220-item benchmark pool via type-stratified random sampling (seed=42) and may share the target misconception label---though not the exact evaluated item---with the test instance. We therefore interpret few-shot results as a format-conditioning baseline rather than a leakage-free estimate of diagnostic generalization.

\paragraph{Generation hyperparameters.} All diagnosis conditions use temperature=0.2, top-$p$=0.95. Max tokens: ZS/FS/RAG=512, CoT=768, GRR generate=256, GRR rerank=512. Random seed: 42. For reasoning models (Qwen3.5-9B), chain-of-thought is suppressed by pre-filling \texttt{</think>} to ensure direct output.


\begin{table}[h]
\centering
\small
\begin{tabular}{llccc}
\toprule
\textbf{Model} & \textbf{Cond.} & \textbf{Before} & \textbf{After} & \textbf{Affected} \\
\midrule
Mistral-7B & ZS & .014 & .077 & 83.6\% \\
Mistral-7B & FS & .073 & .241 & 74.5\% \\
Mistral-7B & CoT & .000 & .214 & 100\% \\
Mistral-7B & RAG & .000 & .232 & 100\% \\
Mistral-7B & GRR & .000 & .255 & 100\% \\
\midrule
Phi-4 & CoT & .423 & .423 & 0\% \\
Phi-4 & RAG & .177 & .400 & 54.5\% \\
Phi-4 & GRR & .241 & .477 & 39.5\% \\
\midrule
Qwen3.5-9B & CoT & .427 & .509 & 16.8\% \\
Qwen3.5-9B & RAG & .268 & .459 & 48.6\% \\
Qwen3.5-9B & GRR & .227 & .518 & 55.9\% \\
\bottomrule
\end{tabular}
\caption{Exp.~2 post-hoc parsing correction: Acc@1 before and after regex ID extraction. ``Affected'' = percentage of outputs requiring correction. The regex \texttt{Ma[A-Z]\textbackslash d\{2,3\}} was applied uniformly. Qwen-2.5-7B and Llama-3.1-8B required no correction.}
\label{tab:parsing-correction}
\end{table}

\section{Human Evaluation Details}
\label{sec:appendix-humaneval-detail}

\paragraph{Sampling pipeline.} The \textbf{clean set} ($n$=100) is drawn from 2,250 Exp.~3 dialogues via two-stage screening: 11 automated heuristic filters retain 1,199 dialogues, then GPT-5.4 qualitative scoring selects 100 with mean quality $\geq$3.0. The \textbf{borderline set} ($n$=100) targets the GPT-5.4 quality 2.0--2.99 range; we exclude answer-leaked dialogues and those with severe role confusion, re-screen the remaining 440 with GPT-5.4 for evaluability (yielding 390), then stratified-sample 100 (20 per model). Both subsets are stratified across models, types, and strategies.

\paragraph{Rubric.} Each dialogue is rated on four dimensions: \textit{correct understanding} (Yes/Partial/No), \textit{reasoning quality} (1--5), \textit{answer revealed} (Yes/No), \textit{dialogue naturalness} (1--5). For comparison with automated criteria, only Yes maps to ``correct''.

\paragraph{Naturalness ratings.} The naturalness dimension (Q4), computed as the per-dialogue median of the three annotators' ratings, averages 3.50/5 (SD 0.67) on the clean subset, 3.33/5 (SD 1.03) on the borderline subset, and 3.42/5 pooled ($n$=200). We report these descriptively only: inter-annotator agreement on this dimension is weak---as expected for a single-item subjective Likert scale---so the values indicate moderate naturalness rather than a validated metric, consistent with simulated dialogues being useful but imperfect proxies. Because all three scoring criteria are applied to the same fixed dialogues, naturalness introduces no unpaired generation confound into the criterion contrast in \S\ref{sec:exp3}, though it bounds how far that contrast's magnitude should be expected to generalize beyond this protocol.

\paragraph{Annotation protocol and tooling.} Annotators were blinded to LLM-judge verdicts. The standalone HTML interface showed the problem context, correct answer, target misconception, model/strategy/type/quality metadata, full tutoring dialogue, optional notes, progress indicator, and export control. The GUI used four rating items: Q1 correct understanding (Yes/Partial/No), Q2 reasoning/explanation quality (1--5), Q3 direct answer revelation by the tutor (Yes/No), and Q4 conversational naturalness (1--5). The Q1 instruction told annotators to judge conceptual understanding in the verification answer rather than numeric match alone. The observed inter-annotator agreement (mean pairwise Cohen's $\kappa$=0.51 on clean, 0.55 on borderline)---which stays moderate across both subsets as expected for subjective four-dimensional ratings---shows that the rubric did not collapse into uniform favorable voting.

\paragraph{Recruitment, compensation, consent, and ethics status.} Annotators met the qualification criteria in \S\ref{sec:human-eval} and were recruited based on mathematics and AI-evaluation experience. Each annotator evaluated 200 items and received a fixed honorarium of US\$0.20 per item, set in proportion to expected annotation time for a low-risk, non-sensitive evaluation task; no external crowdwork platform was used. They were informed that aggregated, anonymized labels and agreement statistics would be used for research reporting. The study used public datasets and model-generated dialogues, involved no real student participants, and collected no sensitive personal data; see also the Ethical Considerations section.

\section{Human--Judge Disagreement Examples}
\label{sec:appendix-disagreements}

Of the 100 clean-set dialogues, the corrected GPT-5-mini judge and human consensus disagree on only 3 cases (2 false positives, 1 false negative). We summarize each to illustrate the boundary of LLM judge reliability.

\paragraph{False positive 1 (HE-034).} Factual misconception (applying $\sqrt{a}\sqrt{b}=\sqrt{ab}$ to negative numbers). The student attempts two approaches but incorrectly applies the real-number product rule in both. The judge credited the student's attempt to compare methods as evidence of understanding; human annotators judged that the core error was not resolved.

\paragraph{False positive 2 (HE-042).} Procedural misconception (choosing $u$ and $dv$ in integration by parts to make the integral harder). The student repeats the same poor choice ($u=\cos x$, $dv=x^3 dx$), producing a more complex integral. The judge credited correct application of the integration-by-parts formula; human annotators judged that the strategic selection error was not remediated.

\paragraph{False negative 1 (HE-081).} Factual misconception (confusing limit existence with continuity). The student correctly computes $\lim_{x\to 0} \sin x/x = 1$ and concludes discontinuity because $g(0)=2 \neq 1$. The judge rejected the response due to an incomplete ``0/0'' argument in the limit computation; human annotators judged that the core concept (limit $\neq$ function value $\Rightarrow$ discontinuous) was correctly applied.

\section{Statistical Test Rationale: McNemar}
\label{sec:appendix-mcnemar}
The strict-vs-judge comparison in \S\ref{sec:exp3} is a paired binary design: each of the 2,250 dialogues is scored under both criteria, so the two RSR figures come from one set of dialogues rather than two independent samples. McNemar's test is the standard test for this design---it tests whether the discordant pairs are symmetric. Here, 949 dialogues are scored as failures by strict match but as passes by the judge, against 67 in the opposite direction, giving $\chi^2 = (949-67)^2/(949+67) = 765.7$. A test that assumed the two criteria had been applied to independent samples would be inappropriate, because the pairing is the point of holding the dialogues fixed; the paired bootstrap CI of [36.6, 41.8]\,pp reported in \S\ref{sec:exp3} corroborates the gap.

\section{TOST Equivalence Testing}
\label{sec:appendix-tost}
We apply the Two One-Sided Tests procedure for two independent proportions \cite{schuirmann1987comparison} to all 15 pairwise condition comparisons in Exp.~4 ($n$=530 per condition: 5 models $\times$ 106 scenarios).  Table~\ref{tab:tost} reports the priority comparisons at $\delta$=5\,pp and $\delta$=10\,pp using the corrected GPT-5-mini judge proportions.

\begin{table}[h]
\centering
\small
\adjustbox{max width=\columnwidth}{%
\begin{tabular}{lcccc}
\toprule
\textbf{Comparison} & $\boldsymbol{\Delta}$ & \textbf{90\% CI} & $\boldsymbol{p_{\text{TOST}}}$ ($\delta$=5) & $\boldsymbol{p_{\text{TOST}}}$ ($\delta$=10) \\
\midrule
C4 vs B1       & +0.6  & [$-$3.6, +4.7] & .039 & $<$.001 \\
C1 vs B1       & +2.1  & [$-$2.0, +6.2] & .119 & $<$.001 \\
C1 vs C4       & +1.5  & [$-$2.5, +5.6] & .078 & $<$.001 \\
C2 vs C4       & $-$2.5 & [$-$6.6, +1.7] & .159 & .002 \\
C3 vs C4       & +2.3  & [$-$1.8, +6.3] & .132 & $<$.001 \\
C3.5 vs C4     & $-$1.1 & [$-$5.3, +3.0] & .063 & $<$.001 \\
\bottomrule
\end{tabular}%
}
\caption{TOST equivalence results for Exp.~4 priority comparisons.  $\Delta$ in percentage points.  At $\delta$=5\,pp only three near-identical pairs are equivalent; at $\delta$=10\,pp all 15 pairwise comparisons are equivalent ($p_{\text{TOST}}<0.05$; only priority pairs shown).}
\label{tab:tost}
\end{table}

\section{Cross-Family Evaluator Validation}
\label{sec:appendix-crossfamily}

Table~\ref{tab:cross-family} provides the full breakdown of the cross-family evaluator validation: every tutor model is scored by all six evaluator models (five open-weight evaluators plus GPT-5.4) under both pipeline and direct conditions, with per-evaluator means reported across the five tutors.

\paragraph{Key finding.} Across the 30 (tutor, evaluator) pairs, evaluator identity systematically affects RSR (Kruskal-Wallis $H$=26.2, $p<0.001$; ICC(2,1)=0.007): four open-weight evaluators inflate RSR (0.41--0.50), while GPT-5.4 (0.26) and Mistral-7B (0.21) are substantially stricter, arguing against same-family collusion as a sufficient explanation. This divergence, however, arises under \emph{mismatched} evaluation context: re-evaluating all 2,250 Exp.~3 dialogues with GPT-5.4 under correctly matched context yields RSR closely tracking GPT-5-mini (78.5\% vs.\ 81.3\%, $\Delta$=2.8\,pp). Two LLM judges thus converge once the evaluation context is aligned, indicating that evaluation context design contributes more to measurement variance than judge identity. The near-zero cross-family ICC initially admits two readings---\textit{evaluator bias} (different standards) or \textit{weak signal} (similar-quality tutoring)---disambiguated by the human evaluation (\S\ref{sec:human-eval}) and the 4-judge convergence in \S\ref{sec:exp3} in favor of evaluator-bias.

\begin{table}[h]
\centering
\small
\adjustbox{max width=\columnwidth}{%
\begin{tabular}{llrccc}
\toprule
\textbf{Evaluator} & \textbf{Family} & \textbf{Params} & \textbf{Pipe.\ RSR} & \textbf{Dir.\ RSR} & \textbf{ARR} \\
\midrule
Qwen3.5-9B & Alibaba & 9B & .50$\pm$.02 & .50$\pm$.02 & .18$\pm$.05 \\
Llama-3.1-8B & Meta & 8B & .45$\pm$.03 & .43$\pm$.03 & .20$\pm$.08 \\
Phi-4 & Microsoft & 14B & .42$\pm$.01 & .42$\pm$.02 & .17$\pm$.02 \\
Qwen-2.5-7B & Alibaba & 7B & .41$\pm$.00 & .41$\pm$.00 & .18$\pm$.04 \\
\midrule
GPT-5.4 & OpenAI & -- & .26$\pm$.02 & .33$\pm$.01 & .16$\pm$.06 \\
Mistral-7B & Mistral AI & 7B & .21$\pm$.03 & .30$\pm$.02 & .15$\pm$.05 \\
\bottomrule
\end{tabular}%
}
\caption{Cross-family evaluator validation (5 tutors $\times$ 6 evaluators, $n$=106 per condition). Mean$\pm$SD across five tutor models. Evaluator identity significantly affects RSR (Kruskal-Wallis $H$=26.2, $p$<0.001; ICC(2,1)=0.007). The horizontal line separates inflating open-weight evaluators from stricter evaluators. ARR averaged over pipeline (C1) and direct (B1) conditions.}
\label{tab:cross-family}
\end{table}

\section{Strategy Fidelity Analysis}
\label{sec:appendix-fidelity}
To verify that strategy-constrained prompts in Exp.~3 actually produced strategy-appropriate tutor behavior, we analyze linguistic markers in all 2,250 tutor dialogues.  For each strategy, we define 7--9 characteristic markers (e.g., Socratic: ``what'', ``why'', ``how'', question marks; Direct Instruction: ``the correct'', ``the rule is'', ``remember that'').  For each dialogue, we count occurrences of the strategy's own markers (\emph{Self}) versus markers belonging to the other four strategies (\emph{Other}), computing the ratio \emph{Self}/(\emph{Self}+\emph{Other}); Table~\ref{tab:fidelity} reports the per-strategy ratios with a chi-squared test.

\begin{table}[h]
\centering
\small
\begin{tabular}{lccc}
\toprule
\textbf{Strategy} & \textbf{Self} & \textbf{Other} & \textbf{Ratio} \\
\midrule
Socratic           & 25.8 & 15.7 & 0.616 \\
Cognitive conflict & 13.9 & 29.7 & 0.325 \\
Analogy            & 10.2 & 36.9 & 0.225 \\
Counterexample     &  4.2 & 37.8 & 0.101 \\
Direct instruction &  2.6 & 24.3 & 0.099 \\
\bottomrule
\end{tabular}
\caption{Strategy fidelity: mean per-dialogue marker counts (\emph{Self} = own strategy markers, \emph{Other} = markers of other strategies).  A $\chi^2$ test on the contingency table confirms strategies are linguistically distinguishable ($\chi^2$=16{,}292, $df$=4, $p<10^{-6}$).}
\label{tab:fidelity}
\end{table}

Socratic questioning achieves the highest self-marker ratio (0.616), as question-based markers (``what'', ``why'', ``?'') dominate its dialogues.  Direct instruction and counterexample have low absolute self-marker counts but are still statistically distinguishable from each other and from the other strategies.  The highly significant $\chi^2$ ($p<10^{-6}$) confirms that strategy-constrained prompts produce meaningfully different linguistic patterns, though the moderate ratios for some strategies indicate substantial overlap in general tutoring language; qualitative examples appear in Appendix~\ref{sec:appendix-strategy-examples}.

\section{Strategy Invariance: Qualitative Examples}
\label{sec:appendix-strategy-examples}

To illustrate how different strategies converge to identical outcomes, we present two example pairs from Qwen-2.5-7B: one where all strategies succeed and one where all fail.

\paragraph{Successful remediation (conceptual-0029).}
This scenario targets confusing independence with mutual exclusivity in probability. All five models across all five strategies successfully remediate this misconception (25/25 cells).

\smallskip\noindent\textbf{Counterexample strategy} (key turns):
\begin{quote}\small
\textbf{Tutor:} ``Imagine rolling a die twice\ldots think about event C, getting a 4 on the first roll. Events A and C are clearly different\ldots However, both A and C can still be independent of event B even though they overlap with B in different ways.''

\textbf{Student:} ``I recalculated: P(A)=1/6, P(B)=1/6, P(A$\cap$B)=1/36. That equals P(A)P(B)\ldots so by the usual definition they are independent.''
\end{quote}

\noindent\textbf{Socratic strategy} (key turns):
\begin{quote}\small
\textbf{Tutor:} ``If A and B are independent, then the occurrence of A shouldn't affect the probability of B\ldots Can you calculate P(A) and P(B) and multiply them?''

\textbf{Student:} ``P(A)P(B)=(1/6)(1/6)=1/36=P(A$\cap$B). That satisfies the independence condition\ldots This contradicts my original conclusion.''
\end{quote}

Both strategies lead the student to the same insight---independence is defined by $P(A \cap B) = P(A)P(B)$, not by whether events share outcomes---but through different pedagogical paths (concrete alternative scenario vs.\ guided calculation).

\paragraph{Failed remediation (procedural-0003).}
This scenario targets a procedural error in multi-step word problems (double-counting adjusted days). All five models fail across all strategies.

\smallskip\noindent\textbf{Counterexample strategy} (key turns):
\begin{quote}\small
\textbf{Tutor:} ``Let's use a counterexample. If Thursday was shortened by 30 minutes instead of 20, and you started with 5 hours (300 minutes), how would you adjust?''

\textbf{Student:} ``5 hours = 300 minutes, then subtract 20 (Thursday) and add 10 (Friday): 300 $-$ 20 $+$ 10 = 290 minutes.''
\end{quote}

\noindent\textbf{Socratic strategy} (key turns):
\begin{quote}\small
\textbf{Tutor:} ``Jeff runs Monday through Friday\ldots how many full days of running at the original one-hour duration are there?''

\textbf{Student:} ``You subtract 20 from Thursday (60$-$20=40) and add 10 on Friday (60$+$10=70). So total = 3$\times$60$+$40$+$70 = 290 minutes.''
\end{quote}

Both students arrive at the correct answer (290) during tutoring but fail the verification problem, misapplying the procedure to a novel scenario. The tutoring provides procedural scaffolding for the specific problem but does not transfer the underlying skill---regardless of strategy.

These examples suggest that scenario difficulty, not strategy, governs the measured outcome: amenable misconceptions yield to any reasonable strategy, while resistant ones withstand all of them.

\section{Persona Decomposition}
\label{sec:appendix-persona}




A natural rebuttal to the small strategy-effect null in \S\ref{sec:exp3} is that the simulated student has a \emph{persuadability ceiling} that flattens any tutoring intervention. To test this alternative, we compare three conditions on all 106 scenarios under LLM-judge scoring:

\begin{itemize}[leftmargin=*,itemsep=2pt,topsep=2pt]
  \item \textbf{(A) No persona}: the student simulator solves the verification problem directly in oracle mode. RSR = 98.1\%.
  \item \textbf{(B) Persona only}: the student adopts the assigned misconception persona but receives \emph{no tutoring} before the verification problem. RSR = 6.6\%.
  \item \textbf{(C) Persona + tutoring}: the standard Exp.~3 setting. RSR = 81.3\%.
\end{itemize}

The 91.5\,pp A$\rightarrow$B drop shows that the simulator adopts the assigned misconception (it is not a yes-machine). The 74.7\,pp B$\rightarrow$C lift shows that tutoring meaningfully moves the score, and the lift is type-dependent (conceptual +63.6\,pp vs.\ procedural +38.0\,pp). Together these results put a floor under simulator richness and argue against a simple persuadability-ceiling explanation for the small strategy effects in the main text---the measured equivalence is not merely a saturation artifact.

\section{Wrong-Diagnosis Control}
\label{sec:appendix-wrong-diag}

To test whether tutors condition on supplied diagnosis, we provide oracle detection with a deliberately wrong same-type misconception label (106 scenarios $\times$ 5 models = 530 dialogues). Wrong labels are selected to be non-duplicate and dissimilar (Jaccard similarity $<$0.7, string-similarity ratio $<$0.8, drawn from a different taxonomy entry). Table~\ref{tab:wrong-diag} reports RSR for the wrong, gold, and direct conditions across three independent judges.

\begin{table}[h]
\centering
\small
\begin{adjustbox}{max width=\columnwidth}
\begin{tabular}{lcccc}
\toprule
\textbf{Condition} & \textbf{GPT} & \textbf{Claude} & \textbf{Gemini} & \textbf{3-judge mean} \\
\midrule
C2: Gold diagnosis & 76.6\% & 71.3\% & 74.5\% & 74.1\% \\
B1: Direct tutoring & 78.5\% & 69.6\% & 73.8\% & 74.0\% \\
C\_wrong: Wrong diagnosis & 80.8\% & 74.5\% & 77.9\% & 77.7\% \\
\midrule
$\Delta$ (Wrong $-$ Gold) & $+$4.2\,pp & $+$3.2\,pp & $+$3.4\,pp & $+$3.6\,pp \\
\bottomrule
\end{tabular}
\end{adjustbox}
\caption{Wrong-diagnosis control results across three independent LLM judges from three vendor families. Column labels abbreviate GPT-5-mini, Claude Haiku 4.5, and Gemini 3.1 Flash Lite. All three judges find $C_{\text{wrong}} \geq C_2$ in the same direction (Wrong $\geq$ Gold by $+$3.2--4.2\,pp), supporting that the negative finding is robust to judge identity, not a GPT-5-mini artifact. With McNemar (GPT-5-mini), neither C\_wrong vs C2 ($p$=.078) nor C\_wrong vs B1 ($p$=.315) reaches significance, indicating that remediation outcomes do not measurably depend on diagnostic correctness under this protocol.}
\label{tab:wrong-diag}
\end{table}

Table~\ref{tab:readings-evidence} summarizes, for each of the four readings of the null (\S\ref{sec:exp4}), which control bears on it and how far; all four remain open. The diagnosis-injection format is single in these experiments---an explicit scope limit: the control tests conditioning under one presentation format, and the claim is narrowed accordingly.

\begin{table}[h]
\centering
\small
\begin{adjustbox}{max width=\columnwidth}
\begin{tabular}{p{0.30\columnwidth}p{0.42\columnwidth}p{0.20\columnwidth}}
\toprule
\textbf{Reading} & \textbf{Bearing evidence} & \textbf{Status} \\
\midrule
(i) Pipeline adds little signal & GLMM null ($p$=0.263); criterion-invariant spread (strict 3.1 vs.\ judge 4.9\,pp) & Consistent; not separated \\
(ii) Tutors do not use the label & 3-judge wrong$\geq$gold direction (+3.2/+3.4/+4.2\,pp) rules out a single-judge artifact & Open; favored jointly with (iii) \\
(iii) Simulator recovers regardless & Persona-only floor 6.6\%; 21.2\,pp profile variance; type-dependent recovery & Bounded, not eliminated \\
(iv) Judge insensitivity & Type separation (89.9/80.9/73.2, $p<10^{-30}$) shows pedagogical sensitivity; swap sensitivity untested & Narrowed, open \\
\bottomrule
\end{tabular}
\end{adjustbox}
\caption{Pipeline-null readings and the controls that bear on each. No control separates readings (ii) and (iii); all four remain open, and the conditioning claim is scoped to the tested diagnosis-presentation format.}
\label{tab:readings-evidence}
\end{table}

The paired bootstrap 95\% CIs include zero (vs.\ C2: [$-$0.2,+8.7]\,pp; vs.\ B1: [$-$1.7,+6.4]\,pp).

\paragraph{Diagnosis-emphasis robustness variant.} As committed during the review discussion, we test whether this null is sensitive to how the diagnosis is presented. We re-run both arms---gold and wrong same-type diagnosis---with exactly two prompt changes: the diagnosed misconception is stated \emph{first}, before the problem and student answer, and the tutor receives an explicit directive to directly address that specific misconception. Everything else (scenarios, profiles, turns, temperatures, simulator, judge) is identical to the control above; both arms are scored fresh by the same judge. This is a robustness check, not support for any claim in the paper.

\begin{table}[h]
\centering
\small
\begin{adjustbox}{max width=\columnwidth}
\begin{tabular}{lccc}
\toprule
\textbf{Tutor} & \textbf{Gold+emph} & \textbf{Wrong+emph} & $\boldsymbol{\Delta}$ \\
\midrule
Qwen-2.5-7B  & 80.2\% & 85.8\% & $+$5.7 \\
Mistral-7B   & 71.7\% & 79.2\% & $+$7.5 \\
Llama-3.1-8B & 88.7\% & 74.5\% & $-$14.2 \\
Phi-4        & 81.1\% & 83.0\% & $+$1.9 \\
Qwen3.5-9B   & 86.8\% & 87.7\% & $+$0.9 \\
\midrule
\textbf{Mean} & \textbf{81.7\%} & \textbf{82.1\%} & $+$\textbf{0.4} \\
\bottomrule
\end{tabular}
\end{adjustbox}
\caption{Diagnosis-emphasis variant: RSR under gold vs.\ wrong same-type diagnosis with emphasized presentation (106 scenarios per cell; 1,060 new dialogues). $\Delta$ = wrong $-$ gold, in percentage points, computed from unrounded counts.}
\label{tab:emphasis-variant}
\end{table}

Even with the diagnosis stated first and an explicit directive to address it, wrong and gold diagnoses remain statistically indistinguishable: 82.1\% vs.\ 81.7\% (Table~\ref{tab:emphasis-variant}; $\Delta$=+0.4\,pp; McNemar $p$=0.93; paired bootstrap 95\% CI [$-$3.8, +4.7]\,pp), with per-model deltas mixed in sign ($-$14.2 to $+$7.5\,pp). The gap is, if anything, smaller than in the original control ($+$4.2\,pp, $p$=0.078). Under this protocol, the null is therefore not an artifact of a weak diagnosis-presentation format; the conditioning claim remains scoped to the tested simulator and tutors. Thus, under this simulated-student protocol, supplied diagnostic labels do not act as a strong causal lever for measured remediation. This does not show that diagnosis is intrinsically irrelevant; rather, it suggests that current 7B--14B tutors may not reliably leverage diagnostic labels, or that generic tutoring and simulator recoverability dominate the effect of diagnostic correctness. Future work with larger or instruction-tuned models may reveal conditions under which correct diagnosis measurably improves outcomes.

\section{Out-of-Family Tutor Replication Details}
\label{sec:appendix-frontier}

The out-of-family replication in \S\ref{sec:exp5} follows Exp.~3 exactly: Claude Haiku 4.5 and Gemini 3.1 Flash Lite as tutors, GPT-5-mini as student simulator (identical profile bank), 5 strategies $\times$ 3 misconception types $\times$ 30 scenarios = 450 dialogues per tutor. Scoring uses the in-experiment GPT-5-mini judge plus post-hoc Claude Haiku 4.5 and Gemini 3.1 Flash Lite judges, reproducing the 3-judge LLM panel from \S\ref{sec:exp3}. Table~\ref{tab:e7} reports per-tutor RSR under each of the three LLM judges plus strict match, alongside the five 7B--14B baselines for direct comparison.

\begin{table}[h]
\centering
\small
\adjustbox{max width=\columnwidth}{%
\begin{tabular}{lcccc}
\toprule
\textbf{Tutor} & \textbf{Strict} & \textbf{GPT} & \textbf{Claude} & \textbf{Gemini} \\
 & \textbf{(\%)} & \textbf{(\%)} & \textbf{(\%)} & \textbf{(\%)} \\
\midrule
\multicolumn{5}{l}{\textit{5 baselines (Exp.~3, recomputed for comparison)}} \\
Qwen-2.5-7B            & 44.4 & 81.8 & 76.2 & 79.1 \\
Mistral-7B             & 40.2 & 77.8 & 70.2 & 73.6 \\
Llama-3.1-8B           & 40.2 & 79.1 & 74.4 & 76.9 \\
Phi-4                  & 42.4 & 83.6 & 78.2 & 81.1 \\
Qwen3.5-9B             & 43.3 & 84.4 & 77.8 & 80.4 \\
\textit{Mean}          & 42.1 & 81.3 & 75.4 & 78.2 \\
\midrule
\multicolumn{5}{l}{\textit{Out-of-family tutors (Exp.~5)}} \\
Claude Haiku 4.5              & 43.1 & 87.8 & 81.6 & 85.3 \\
Gemini 3.1 Flash Lite      & 43.1 & 86.2 & 78.7 & 84.0 \\
\midrule
$\Delta$ (out-of-family mean $-$ baseline) & $+$1.0 & $+$5.7 & $+$4.8 & $+$6.5 \\
\bottomrule
\end{tabular}%
}
\caption{Exp.~5: RSR (\%) for out-of-family tutors under each LLM judge, with the five 7B--14B baselines from Exp.~3 for comparison. Column labels abbreviate GPT-5-mini, Claude Haiku 4.5, and Gemini 3.1 Flash Lite.}
\label{tab:e7}
\end{table}

By misconception type (GPT-5-mini judge): Claude Haiku 4.5 reports conceptual 94.0\%, factual 85.3\%, procedural 84.0\%; Gemini 3.1 Flash Lite reports conceptual 91.3\%, procedural 87.3\%, factual 80.0\%. The conceptual-highest pattern from the baselines (\S\ref{sec:exp3}) is preserved.

By strategy (GPT-5-mini judge): Claude Haiku 4.5 ranges 82.2\%--93.3\% (spread 11.1\,pp, analogy highest); Gemini 3.1 Flash Lite ranges 81.1\%--90.0\% (spread 8.9\,pp, analogy highest). Both are slightly wider than the baseline 6.7\,pp range but remain narrow, consistent with the flat-strategy finding in \S\ref{sec:exp3}.

Answer Reveal Rate: Claude Haiku 4.5 reports 8.9\% and Gemini 3.1 Flash Lite reports 7.1\%---near the baseline ARR range (6.0--8.7\%), with Claude Haiku 4.5 just 0.2\,pp above it.

\section{Reproducibility}
\label{sec:appendix-reproducibility}

\paragraph{Software.} Python 3.11, Transformers 5.3.0, PyTorch 2.6.0. GPT-5-mini accessed via OpenAI API (March 2026).

\paragraph{Hardware.} All local model experiments run on NVIDIA H100 80GB GPUs. Single-GPU inference for 7B--9B models; dual-GPU for Phi-4 (14B) when co-loading tutor and evaluator.

\paragraph{Model identifiers.}
\begin{itemize}[nosep,leftmargin=*,labelsep=0.35em]
\item Qwen-2.5-7B: \path{Qwen/Qwen2.5-7B-Instruct}
\item Mistral-7B: \path{mistralai/Mistral-7B-Instruct-v0.3}
\item Llama-3.1-8B: \path{meta-llama/Llama-3.1-8B-Instruct}
\item Phi-4: \path{microsoft/phi-4}
\item Qwen3.5-9B: \path{Qwen/Qwen3.5-9B}
\item Claude Haiku 4.5 (out-of-family tutor / judge): \path{anthropic/claude-haiku-4.5} via OpenRouter
\item Gemini 3.1 Flash Lite (out-of-family tutor / judge): \path{google/gemini-3.1-flash-lite} via OpenRouter
\end{itemize}

\paragraph{Reasoning models.} Qwen3.5-9B produces \texttt{<think>...</think>} blocks before its answers. We strip these blocks before evaluation. For this model, minimum generation length is set to 2,048 tokens to accommodate the reasoning trace.

\end{document}